\documentclass[lettersize,journal]{IEEEtran}
\usepackage[implicit=false]{hyperref} 
\usepackage{amsmath,amsfonts}
\usepackage{algorithmic}
\usepackage{algorithm}
\usepackage{array}
\usepackage{xcolor}
\usepackage[caption=false,font=normalsize,labelfont=sf,textfont=sf]{subfig}
\usepackage{textcomp}
\usepackage{stfloats}
\usepackage{url}
\usepackage{verbatim}
\usepackage{graphicx}
\usepackage{cite}
\usepackage{multirow}
\usepackage{amsmath}
\usepackage{booktabs}
\usepackage{txfonts}
\usepackage{bbding}
\usepackage{pifont}
\begin{document}

\title{AeroVerse: UAV-Agent Benchmark Suite for Simulating, Pre-training, Finetuning, and Evaluating Aerospace Embodied World Models}

 \author{{Fanglong~Yao}$^\dag$\href{https://orcid.org/0000-0003-4187-9755}{\includegraphics[scale=0.07]{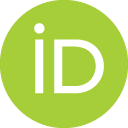}}, \IEEEmembership{Member,~IEEE,} {Yuanchang~Yue}$^\dag$, {Youzhi~Liu}$^\dag$, Xian Sun\href{https://orcid.org/0000-0002-0038-9816}{\includegraphics[scale=0.07]{figures/ORCIDiD_icon.png}}, \IEEEmembership{Senior~Member,~IEEE,} Kun~Fu, \IEEEmembership{Senior~Member,~IEEE}% <-this % stops a space
   
\thanks{This work is supported by the National Natural Science Foundation of China under Grants 62306302 and 62425115. \textit{(Corresponding author: Fanglong~Yao.)} Authors labeled with {$\dag$} contribute equally.} 

\thanks{Fanglong~Yao is with the Aerospace Information Research Institute, Chinese Academy of Sciences, Beijing 100190, China, and with the Key Laboratory of Target Cognition and Application Technology(TCAT), Aerospace Information Research Institute, Chinese Academy of Sciences, Beijing 100190, China (e-mail: yaofanglong17@mails.ucas.ac.cn).}% <-this % stops a space
    \thanks{Yuanchang Yue,~Youzhi Liu,~Xian Sun,~Kun Fu are with the Aerospace Information Research Institute, Chinese Academy of Sciences, Beijing 100190, China, and also with the University of Chinese Academy of Sciences, Beijing 100190, China, and with the School of Electronic, Electrical and Communication Engineering, University of Chinese Academy of Sciences, Beijing 100190, China, and with the Key Laboratory of Target Cognition and Application Technology(TCAT), Aerospace Information Research Institute, Chinese Academy of Sciences, Beijing 100190, China (e-mail:
    yueyuanchang22@mails.ucas.ac.cn;
   liuyouzhi22@mails.ucas.ac.cn;
    sunxian@aircas.ac.cn; kunfuiecas@gmail.com).
    }% <-this % stops a space
% \thanks{Zhigang Wang is with Shanghai AI Laboratory, (e-mail: wangzhigang@pjlab.org.cn)}
% \thanks{Bin Zhao is with the School of Artificial Intelligence, OPtics and ElectroNics (iOPEN), Northwestern Polytechnical University, (e-mail: binzhao111@gmail.com)}
% \thanks{Jian Zhao is with the Institute of AI (TeleAI), China Telecom and with the School of Artificial Intelligence, Optics and Electronics (iOPEN), Northwestern Polytechnical University (NWPU), Xi’an China, (e-mail: zhaoj90@chinatelecom.cn)}
% \thanks{Lei Jin is currently an Associate Research Fellow with the Beijing University of Posts and Telecommunications (BUPT), Beijing, China, (e-mail: jinlei@bupt.edu.cn )}
	}

\markboth{Journal of \LaTeX\ Class Files,~Vol.~14, No.~8, August~2021}%
{Shell \MakeLowercase{\textit{et al.}}: A Sample Article Using IEEEtran.cls for IEEE Journals}


\maketitle

\begin{abstract}
Aerospace embodied intelligence aims to empower unmanned aerial vehicles (UAVs) and other aerospace platforms to achieve autonomous perception, cognition, and action, as well as egocentric active interaction with humans and the environment. The aerospace embodied world model serves as an effective means to realize the autonomous intelligence of UAVs and represents a necessary pathway toward aerospace embodied intelligence. [\textit{Background}] However, existing embodied world models primarily focus on ground-level intelligent agents in indoor scenarios, while research on UAV intelligent agents remains unexplored, lacking systematic and standardized benchmark suites. [\textit{Aim}] To address this gap, this study aims to construct a comprehensive benchmark suite, AeroVerse, to facilitate the simulation, pre-training, finetuning, and evaluation of aerospace embodied world models. [\textit{Innovations}] We develop AeroSimulator, a simulation platform that encompasses four realistic urban scenes for UAV flight simulation. Additionally, we construct the first large-scale real-world image-text pre-training dataset from a first-person UAV perspective, AerialAgent-Ego15k, and create a virtual image-text-pose alignment dataset, CyberAgent-Ego500k, to facilitate the pre-training of the aerospace embodied world model. We clearly define five downstream tasks for the first time, i.e., aerospace embodied scene awareness, spatial reasoning, navigational exploration, task planning, and motion decision, and have constructed corresponding instruction datasets for fine-tuning. We also develop SkyAgent-Eval, a downstream task evaluation system based on GPT-4. Furthermore, we propose SkyAgentX, the first UAV-agent large model integrating “perception-reasoning-navigating-planning”, which innovatively incorporates aerospace embodied chain-of-thought mechanism and multitask curriculum learning strategy. [\textit{Results}] By benchmarking ten mainstream models, our results reveal the significant limitations of existing 2D/3D visual-language models in complex aerospace embodied tasks and demonstrate the superior performance of SkyAgentX, which outperforms existing methods by an average of 8.52\% across four core tasks, underscoring the necessity and contribution of our work. The AeroVerse benchmark suite will be released to the community to promote exploration and development of aerospace embodied intelligence. \textit{(https://github.com/06f081zyd/AeroVerse)}
\end{abstract}


\section{Introduction}
\IEEEPARstart{D}{rones} have a wide range of applications, including mountainous photovoltaic inspection, river trash detection, pedestrian traffic monitoring at intersections, electric power inspection, and forest fire rescue \cite{Dji1}. However, these applications often depend on manual remote control of the drones. For instance, in UAV mountain photovoltaic inspections, it is necessary to deploy professional operators who spend several hours each day inspecting multiple stations. This practice can easily lead to operator fatigue, resulting in component defects and missed inspections. Therefore, there is an urgent need for UAVs equipped with autonomous intelligence to reduce costs and enhance efficiency.

Aerospace embodied intelligence refers to the specialized application of embodied intelligence within the aerospace sector, focusing on empowering unmanned platforms such as satellites, drones, and aircraft to autonomously integrate perception, cognition, and action. This integration aims to facilitate egocentric active interactions with both humans and the physical environment. Over the past year, visual-language models that encode world knowledge have rapidly advanced, driven by a wealth of high-fidelity simulators and datasets\cite{Shridhar2019ALFREDAB,ku-etal-2020-room,Li2021iGibson2O,Qi2019REVERIERE,Shen2020iGibson1A,Xia2019InteractiveGB,gao2025openfly}, thereby presenting new opportunities for embodied intelligence. Numerous embodied world models\cite{huang2023embodied,Hong20233DLLMIT,Driess2023PaLMEAE,Brohan2023RT2VM,Mu2023EmbodiedGPTVP,X2-VLM,EnhancingVisualGrounding,Vision-LanguageModelsASurvey}, have emerged, significantly enhancing the capabilities of embodied agents in perceiving their surroundings and planning tasks. Consequently, this article posits that the development of an aerospace embodied world model is a crucial strategy for achieving autonomous intelligent agents for drones and represents a necessary pathway toward advancing aerospace embodied intelligence.
\renewcommand{\dblfloatpagefraction}{0.9}
\begin{figure*}[t]
	\centering
		\includegraphics[scale=.27]{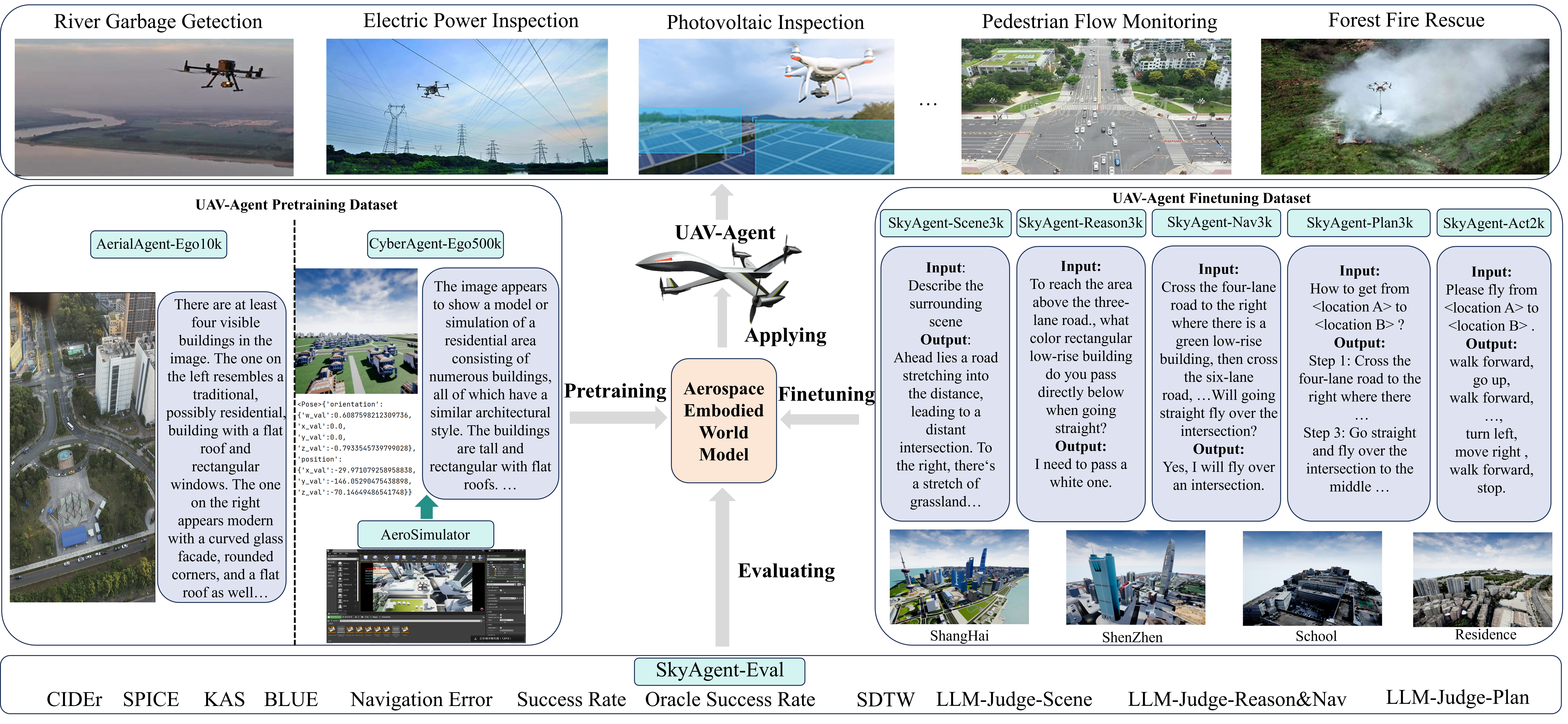}
	\caption{The benchmark suite for the aerospace embodiment world model, AeroVerse, comprises one simulation platform (AeroSimulator), two real-virtual pre-training datasets (AerialAgent-Ego15k and CyberAgent-Ego500k), five downstream task instruction datasets (SkyAgent-Scene3k, SkyAgent-Reason3k, SkyAgent-Nav3k and SkyAgent-Plan3k, and SkyAgent-Act3k
 ), and more than ten evaluation metrics (SkyAgent-Eval). 
 }
	\label{FIG:framework}
\end{figure*}

However, these embodied world models primarily focus on indoor scenarios (e.g., robotic arms) or ground-based agents in outdoor environments (e.g., unmanned vehicles) \cite{huang2023embodied, Hong20233DLLMIT, Driess2023PaLMEAE, Brohan2023RT2VM, Mu2023EmbodiedGPTVP}. There has been limited exploration of UAV embodied agents, particularly in the context of aerospace embodied world models that facilitate UAV autonomy, which is significantly constrained by the development of UAV embodied datasets. In contrast to indoor embodied intelligence datasets \cite{Hong20233DLLMIT, huang2023embodied}, several key challenges arise in the construction of UAV embodied intelligence datasets:

\textbf{Lack of Definition of UAV Embodied Tasks}. In recent years, research on ground-oriented agents has gained significant attention, leading to clearer definitions of downstream tasks such as indoor/outdoor navigation\cite{ku-etal-2020-room,Qi2019REVERIERE}, command following\cite{Shridhar2019ALFREDAB}, and embodied question answering. However, UAV agents must comprehend the intrinsic correlations of four-dimensional space-time and perform actions under conditions of scene randomization and local observability of the environment. This involves aspects such as awareness, cognition, planning, and decision-making. The diversity and interdependence of these downstream tasks result in a lack of clarity in the task definitions for aerial-embodied agents.

\textbf{Difficulty in UAV 3D Data Acquisition.} The widespread use of LiDAR technology in mobile smart devices has facilitated the easy acquisition of indoor 3D data, leading to substantial accumulation. In contrast, obtaining outdoor 3D data necessitates specialized equipment, such as drones, which presents a higher barrier to entry. Furthermore, outdoor 3D data acquisition requires skilled professionals to operate drones and collect extensive point cloud data over larger areas.

\textbf{High Cost of UAV Embodied Data Collection}. UAVs possess a greater range of motion compared to ground agents (e.g., indoor sweeping robots), allowing for a high degree of freedom in three-dimensional space. They can operate over extensive areas (ranging from dozens to hundreds of square kilometers) and navigate complex environments characterized by irregularly distributed obstacles, e.g., buildings and trees. Consequently, this necessitates extensive training for annotators to effectively conduct data collection for UAV agents.

Therefore, our paper, for the first time, explicitly defines five downstream tasks for UAV-embodied agents, highlighting directions for further exploration in this field, as follows:

\begin{itemize}
    \item Aerospace Embodied Scene Awareness: UAV-agent perceives the surrounding 3D environment from a first-person perspective to enhance scene understanding.
    \item Aerospace Embodied Spatial Reasoning: The UAV agent models the spatial relationships between objects within a 3D scene, enabling reasoning about the relationships among these objects.
    \item Aerospace Embodied Navigational Exploration: The UAV agent comprehends navigation commands and navigates to the destination while describing the environment.
    \item Aerospace Embodied Task Planning: UAV-agent generates detailed, landmark-level long-range path planning scenarios to reach the destination.
    \item Aerospace Embodied Motion Decision: The UAV agent provides a complete sequence of actions from the starting point to the destination, thereby realizing an end-to-end closed loop of the scene awareness, path planning, and action decision-making.
\end{itemize}

As illustrated in Figure \ref{FIG:framework}, we address the gap in the UAV-agent dataset and enhance the training of aerospace embodied world models by constructing the first large-scale virtual-reality pre-training dataset alongside a high-quality instruction dataset. Specifically, the first-person, high-resolution real-world pre-training dataset of  high-altitude drones, AerialAgent-Ego15k, is derived from the UrbanBIS dataset. Additionally, we develop the aligned pre-training dataset, CyberAgent-Ego500k, which includes perspective images, scene text descriptions, and drone attitudes. Furthermore, we create five downstream task instruction datasets: SkyAgent-Scene3k, SkyAgent-Reason3k, SkyAgent-Nav3k and SkyAgent-Plan3k, and SkyAgent-Act3k. These datasets are constructed using our established simulation platform, AeroSimulator, which employs Unreal Engine 4, the Microsoft AirSim drone simulator\cite{airsim2017fsr}, and the 3D UrbanScene virtual city dataset\cite{Liu2021UrbanScene3DAL}. This encompasses four real-world urban scenarios, i.e., Shanghai, Shenzhen, School, and Residence, with a coverage area ranging from $1.0\times10^{5}$ to $3.7\times10^{7} m^{2}$. AerialAgent-Ego15k is utilized to enhance the model’s ability to comprehend real urban scenes. CyberAgent-Ego500k is designed for virtual alignment pre-training of visual-language-posture for the Aerospace Embodied World Model, aiming to improve the model’s fundamental generalization capacity in simulated environments. This dataset contains 500K aligned UAV postures, first-person view images, and text descriptions, collected from four 3D urban environments. The collection principle prioritizes images featuring complex scenes, including buildings, roads, and trees. Furthermore, the downstream task instruction datasets are compiled by ten trained professional annotators who operated the UAV in a 3D urban environment for data acquisition and annotation. This process took a total of eight months and underwent rigorous quality checks to ensure the accuracy and reliability of the data, making it ideal for fine-tuning and evaluating the performance of downstream tasks.

Furthermore, we develop a range of scientific automated evaluations, i.e., SkyAgent-Eval, for downstream tasks. Previous advancements have introduced various rubrics for text generation tasks, including BLUE \cite{Papineni2002BleuAM}, SPICE\cite{Anderson2016SPICESP}. 
% \cite{lin-2004-rouge}, CIDEr \cite{Vedantam2014CIDErCI}, and BERTScore \cite{Zhang2019BERTScoreET}
These methods assess text quality from relatively fixed and limited perspectives, such as semantic similarity and word matching, which can impede their customization and adaptability for evaluating downstream tasks involving UAV agents. Moreover, most existing methods depend on probabilistic statistics and do not align with human preferences. In contrast, large language models \cite{Achiam2023GPT4TR,Touvron2023LLaMAOA,Bai2023QwenTR}, trained using reinforcement learning with human feedback (RLHF)\cite{lambert2022illustrating}, generate responses that more accurately reflect human values and preferences. This makes them a viable alternative for evaluating text generation while mitigating the high costs associated with human evaluation\cite{Zheng2023JudgingLW}. Therefore, by leveraging the multifaceted capabilities of large language models, we propose an automated evaluation approach based on GPT-4 \cite{Achiam2023GPT4TR} for three types of downstream tasks, specifically LLM-Judge-Scene, LLM-Judge-Reason$\&$Nav, and LLM-Judge-Plan. This approach employs few-shot instruction and context learning to cater to the customized evaluation needs of various downstream tasks, thus facilitating a more comprehensive and objective assessment of their performance.

Based on the high-quality datasets and evaluation metrics, we propose SkyAgentX, the first UAV-agent embodied large model integrating “perception-reasoning-navigating-planning”. Through the innovative introduction of the aerospace embodied chain-of-thought mechanism, SkyAgentX achieves closed-loop collaboration of environmental perception, spatial relationship reasoning, and multi-step task planning. Meanwhile, combined with a multitask curriculum learning strategy, the model can progressively adapt to the demands of dynamic scenarios from simple to complex. Experiments demonstrate that SkyAgentX outperforms existing methods by an average of 8.52\% in four core tasks (scene perception, spatial reasoning, navigation exploration, and task planning), providing a generalized solution for autonomous drone intelligence systems and opening a new research direction for aerospace embodied intelligence.

In summary, The contributions can be summarized as follows:

(1) We construct the first large-scale real-world image-text pre-training dataset, AerialAgent-Ego15k, utilizing urban UAVs as the primary viewpoint. Additionally, we develop the virtual image-text-posture alignment dataset, CyberAgent-Ego500k, to pre-train the aerospace embodied world model, thereby enhancing the UAV agent’s ability to adapt to both real and virtual environments.

(2) For the first time, we clearly define five aerospace embodied downstream tasks: scene awareness, spatial reasoning, navigational exploration, task planning, and motion decision-making. To support the fine-tuning of the aerospace embodied world model, we create five corresponding instruction datasets, i.e., SkyAgent-Scene3k, SkyAgent-Reason3k, SkyAgent-Nav3k and SkyAgent-Plan3k, and SkyAgent-Act3k, which facilitates the realization of an end-to-end closed-loop of perception, cognition, and action for UAV agents.

(3) We develop a series of automated evaluation methods, i.e., SkyAgent-Eval, based on GPT-4 for the downstream tasks. These methods assess the results comprehensively, flexibly, and objectively, providing quantitative scores and corresponding explanations for task evaluations, thereby enhancing the credibility of the evaluation outcomes.

(4) SkyAgentX pioneers as the first UAV-agent embodied large model with its end-to-end ``perception-reasoning-navigating-planning" framework. This is achieved by integrating aerospace embodied chain-of-thought and multitask curriculum learning, resulting in comprehensive superiority over existing methods—marked by an average performance improvement of 8.52\% across four core tasks.

(5) Extensive experiments are conducted using ten mainstream baselines to analyze their performance on the downstream instruction datasets. The experimental results reveal the potential and limitations of 2D$\//$3D visual-language models in UAV-agent tasks, underscoring the necessity of constructing an aerospace embodied world model.

(6) We design over $10$ 2D$\//$3D visual-language models, $2$ pre-training datasets, $5$ downstream task instruction datasets, and $10+$ evaluation metrics, as well as a simulator featuring $4$ urban scenarios, into a benchmark suite, AeroVerse, which will be released to the community to advance the field of aerospace embodied agents.

\section{Related Work}

\subsection{3D Visual-Language Datasets}

The three-dimensional (3D) world encompasses not only horizontal and vertical dimensions but also depth, providing richer information than two-dimensional (2D) images. Depth accurately reflects fundamental aspects of the real world and enhances the ability of embodied agents to learn from and understand their 3D environment. Furthermore, textual annotations accompanying 3D visual-language datasets assist embodied agents in perceiving their surroundings and conducting spatial reasoning. However, challenges in creating 3D datasets have led to a scarcity of such resources, with only a limited number of datasets publicly available to date. For instance, the ScanQA dataset comprises $41,363$ unique Q\&A pairs, accompanied by 3D object localization annotations for $800$ indoor 3D scenes \cite{2021ScanQA}. The ScanRefer dataset contains $11,046$ distinct Q\&A pairs for $1,613$ indoor 3D scenes \cite{2020ScanRefer}. The ScanNet dataset includes $1,513$ indoor scenes featuring a total of $21$ object categories \cite{J2021ScanNet}. %Additionally, the HM3D dataset provides 3D reconstruction data for $1,000$ buildings \cite{2021Habitat}.

In contrast to the aforementioned 3D visual-language datasets that focus on indoor environments, we have pioneered the development of a constructed 3D dataset that emphasizes large-scale urban scenes. This dataset encompasses areas ranging from ($1.0 \times 10^5$) to ($3.7 \times 10^7$) square meters and includes four representative urban environments, i.e., Shenzhen, Shanghai, Residence, and School. We select flying vehicles, specifically unmanned aerial vehicles (UAVs), as the agents due to their greater degree of freedom.

\subsection{Embodied Intelligence Datasets}
The embodied world model serves as an effective approach for empowering embodied agents to interact with their environments, autonomously plan, make decisions, act, and perform tasks similar to human capabilities. Most existing embodied world models concentrate on mobile robots in indoor settings. For example, in the embodied question-and-answer task, Abhishek et al. introduce the EQA dataset, which consists of $9,000$ question-and-answer pairs across $774$ indoor rooms \cite{2018Embodied}. In the domain of embodied task planning, Mohit et al. present the ALFRED dataset, which includes $25,743$ commands and $428,322$ image-action pairs \cite{2019ALFRED}. Additionally, in the realm of embodied navigation tasks, Anderson et al. propose the R2R dataset, which comprises $21,567$ navigation instructions with an average length of $29$ words \cite{2018Vision}.

In contrast, we explicitly define five types of embodied downstream tasks for UAV agents for the first time, each characterized by distinctive features. Taking embodied navigational exploration as an example, we require the agent not only to follow instructions to navigate to a designated destination but also to describe object attributes, such as the color, shape, and height of the building’s floors. Furthermore, we construct five instruction datasets for downstream tasks, namely SkyAgent-Scene3k, SkyAgent-Reason3k, SkyAgent-Nav3k and SkyAgent-Plan3k, and SkyAgent-Act3k. Additionally, we establish a 3D urban simulator, i.e., AeroSimulator, for training UAV agents and collecting data, significantly narrowing the gap between the agents and the real physical environment, thereby facilitating a smoother transition to real-world scenarios.

%\subsection{Language generation class task evaluation metrics} Currently, the traditional evaluation metric in natural language processing tasks is the BLEU metric proposed by Kishore Papineni et al \cite{2002BLEU}, which rates models by comparing the similarity between model outputs and reference texts. It is calculated based on n-gram matching as well as phrase-level exact matching, and is categorized as BLUE-1, BLUE-2, BLUE-3, and BLUE-4 based on the size of the n-grams.In addition, there is also the ROUGE metric that measures the quality of the model outputs by comparing the degree of overlap between model-generated text and the reference text \cite{2004ROUGE}. However, all of the above metrics are based on the matching of text in the reference text and are not as flexible as they are when evaluated by humans. In contrast, our proposed GPT4-based evaluation system can not only address the limitations of traditional metrics, but also be more efficient and less costly.%

\section{Task Formulation}
\renewcommand{\dblfloatpagefraction}{0.9}
\begin{figure*}[t]
	\centering
		\includegraphics[scale=.38]{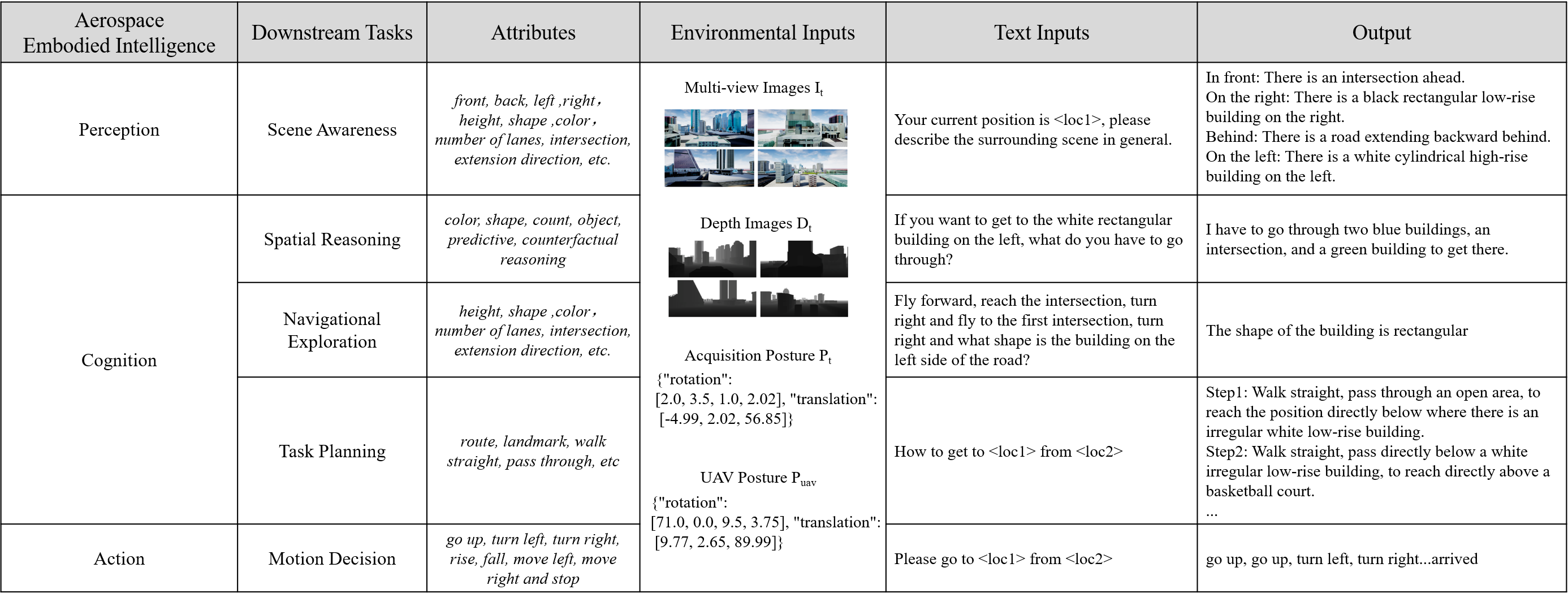}
	\caption{Clear definitions of the five downstream tasks related to aerospace embodied intelligence encompass all aspects of UAV perception, cognition, and action in an end-to-end manner. 
 }
	\label{FIG:task_formulation}
\end{figure*}

\renewcommand{\dblfloatpagefraction}{0.9}
\begin{figure*}[t]
	\centering
		\includegraphics[scale=.39]{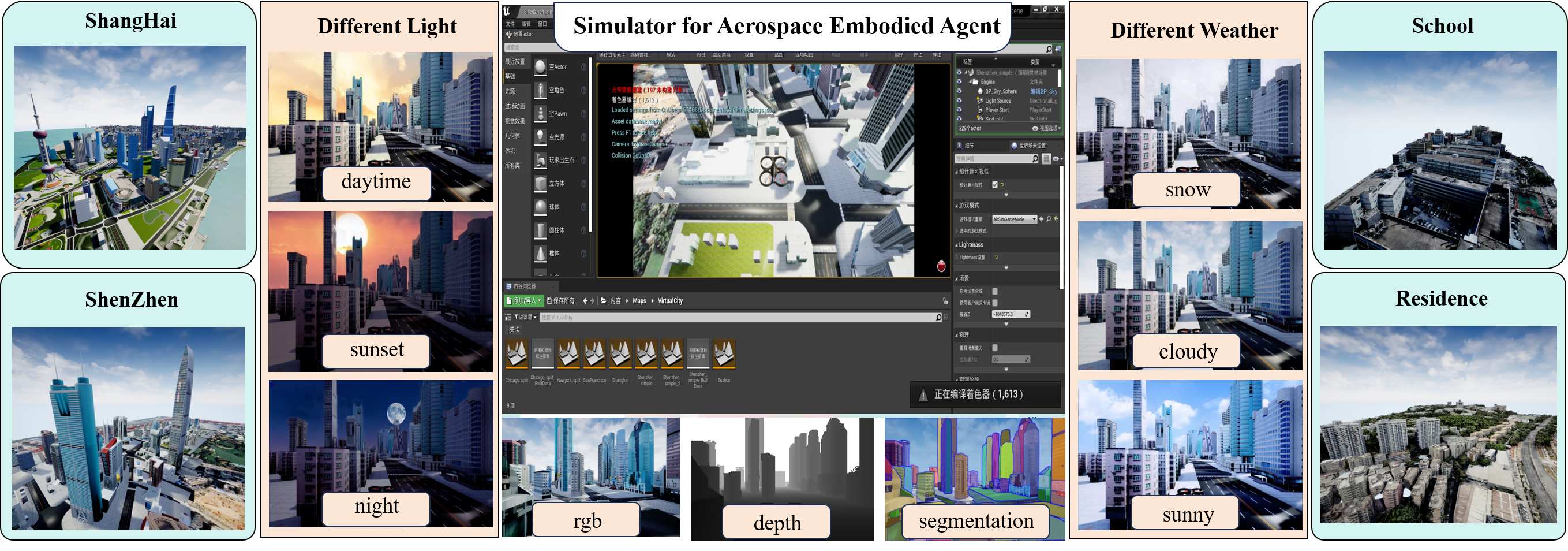}
	\caption{Following the principle of real-to-sim-to-real, we have developed a simulator called AeroSimulator for aerospace embodied agents, such as UAVs. This simulator features four realistic urban environments: Shanghai, Shenzhen, a school, and a residential area. It is capable of simulating various lighting conditions and weather scenarios while generating visual outputs, including RGB images, depth maps, and segmentation data. This functionality significantly reduces the disparity between simulated environments and the real physical world. 
 }
	\label{FIG:simulator}
\end{figure*}

To facilitate the closed-loop training of perception, cognition, and action in UAV agents and to endow them with autonomous capabilities, this paper categorizes the downstream tasks into five distinct categories, as illustrated in Figure \ref{FIG:task_formulation}. It clearly defines the concepts associated with these tasks, standardizes the input and output formats, and offers innovative perspectives for further research on aerospace embodied intelligence in the context of UAVs.

\textbf{Aerospace Embodied Scene Awareness}. Given the current state of drone intelligent agents, specifically their position in three-dimensional space, drones describe surrounding environmental elements, such as buildings, in a panoramic manner (covering four directions: front, back, left, and right). This capability is essential for the cognitive processes and actions of intelligent agents. Traditional environmental perception tasks generally involve inputting environmental images, extracting features from these images, and generating corresponding descriptions. In contrast, the objective of this task is to enhance the ability of UAV agents to perceive their environment and articulate 3D scenes based on their location coordinates.

\textit{Input}: Multi-perspective 2D images of the city's 3D scene, including $I_t=\{i_{t, k}\}_{k=1}^K$, depth map 
$D_t=\{d_{t, k}\}_{k=1}^K$, multi-perspective camera pose $P_t=\{p_{t, k}\}_{k=1}^K$, and the current attitude of the drone in the environment.

\textit{Output}: Scene element description $TEXT_{surrounding}$ of UAV agent in four directions, i.e., front, back, left, right,
\begin{equation*}
TEXT_{surrounding}=f(I_t,D_t,P_t,p_{uav},TEXT_{question})
\end{equation*}

\textbf{Aerospace Embodied Spatial Reasoning}. Based on the current location and three-dimensional environment, the drone agent infers the object’s orientation relationships, action trajectories, and counterfactual scenarios within the scene, guided by specific questions. The objective is to enhance the agent’s understanding of the 3D spatial scene graph, which is a fundamental task of embodied cognition. Traditional spatial reasoning tasks primarily focus on recognizing spatial relationships between objects in a single 2D image, characterized by simplistic scenes and a limited number of objects. In contrast, this task emphasizes reasoning about relationships, intentions, counterfactuals, and other dimensions within three-dimensional space, which is inherently more complex and aligns more closely with human logical reasoning.

\textit{Input}: Multi-perspective 2D images of urban 3D scenes, including $I_t=\{i_{t, k}\}_{k=1}^K$, depth map $D_t=\{d_{t, k}\}_{k=1}^K$, multi-perspective camera pose  $P_t=\{p_{t, k}\}_{k=1}^K$, current drone pose $p_{uav}$ in the environment, question $TEXT_{question}$.

\textit{Output}: The answer $TEXT_{answer}$ to the question, i.e.,
\begin{equation*}
TEXT_{answer}=f(I_t,D_t,P_t,p_{uav},TEXT_{question})
\end{equation*}

\textbf{Aerospace Embodied Navigational Exploration}. Given the UAV agent’s initial position and its long-range, multi-stage navigation instructions, the agent is required to autonomously explore a large urban environment and answer questions related to object characteristics, such as the shape and color of buildings. This capability directly supports applications like object search and tracking in urban settings where building obstructions exist. Unlike traditional navigation tasks that rely solely on navigation instructions and do not include question-answering functions, this task necessitates that the agent not only autonomously navigate and explore its surroundings according to the provided instructions but also respond to inquiries based on the information it collects.

\textit{Input}: Multi-perspective 2D images of urban 3D scenes with $I_t=\{i_{t, k}\}_{k=1}^K$, depth map $D_t=\{d_{t, k}\}_{k=1}^K$, multi-perspective camera pose $P_t=\{p_{t, k}\}_{k=1}^K$, current drone pose in the environment $p_{uav}$, navigation command $TEXT_{nav}$, $TEXT_{question}$.

\textit{Output}: The answer $TEXT_{answer}$ to the question, i.e.,
\begin{equation*}
TEXT_{answer}=f(I_t,D_t,P_t,p_{uav},TEXT_{nav},TEXT_{question})
\end{equation*}

\textbf{Aerospace Embodied Task Planning}. By specifying the initial position and the anticipated endpoint for the UAV intelligent agent, the agent integrates the 3D environment to generate a detailed, step-by-step path planning process. This process requires the identification of distinct landmarks at each stage, which serves as  the core task in UAV embodied cognition. Current path planning methods for indoor environments primarily focus on coarse-grained paths within a single room. In contrast, this task addresses large-scale urban scenes, where the starting and ending points may be separated by several city blocks. During maneuvers such as turning, moving straight, and ascending, the agent will identify observable landmark-level objects to enhance the accuracy of the path planning.

\textit{Input}: Multi-perspective 2D images of urban 3D scenes, including $I_t=\{i_{t, k}\}_{k=1}^K$, depth map $D_t=\{d_{t, k}\}_{k=1}^K$, multi-perspective camera pose $P_t=\{p_{t, k}\}_{k=1}^K$, as well as the current attitude of the drone, $p_{uav}$, and target pose $p_{end}$.

\textit{Output}: Step-by-step path plans $TEXT_{plan}$ and intermediate pose $p_{temp}$ , i.e.,
\begin{equation*}
TEXT_{plan},p_{temp}=f(I_t,D_t,P_t,p_{uav},p_{end})
\end{equation*}

\textbf{Aerospace Embodied Motion Decision}. The intelligent drone agent operates in real-time, guided by its initial position and target endpoint. It dynamically interacts with its environment and adjusts its action strategy based on the outcomes of each movement and the historical sequence of actions. This iterative process continues until it reaches the endpoint. Unlike traditional decision-making tasks, this approach positions the drone as the agent, making decisions informed by first-person environmental observations at each navigation node. It encompasses a nearly complete end-to-end closed-loop of task chains, including perception, reasoning, planning, and action, representing the ultimate objective for drone agents.

\textit{Input}: Multi-perspective 2D images of urban 3D scenes, including $I_t=\{i_{t, k}\}_{k=1}^K$, depth map $D_t=\{d_{t, k}\}_{k=1}^K$, multi-perspective camera pose $P_t=\{p_{t, k}\}_{k=1}^K$, position $P_{history}=\{p_n\}_{n=0}^{N-1}$, $I_{history}=\{i_n\}_{n=0}^{N-1}$, $A_{history}=\{a_n\}_{n=0}^{N-1}$ from $0$ to $N$-$1$, and target pose $p_{end}$.

\textit{Output}: Action $a_N$ at time $N$, i.e.,
\begin{equation*}
a_{N}=f(I_t,D_t,P_{history},I_{history},A_{history},p_{end})
\end{equation*}

\section{Simulation Platform}
\textbf{Simulator}. To simulate a realistic drone flight scenario, we utilize Unreal Engine $4$ to load urban environments and select AirSim\cite{airsim2017fsr} for constructing the drone model. This enables us to develop a simulator, named AeroSimulator, capable of facilitating multiple action spaces for the drone, as illustrated in Figure \ref{FIG:simulator}. Adhering to the real-to-sim-to-real paradigm, we select four representative scenes from the high-quality UrbanScene3D dataset\cite{Liu2021UrbanScene3DAL} created by Lin et al.: Shenzhen, Shanghai, School, and Residence, all derived from 3D reconstructions of actual physical locations. Furthermore, the simulator accommodates various lighting conditions (day, evening, night, etc.), seasonal variations (spring, summer, autumn, winter), and climatic modes (sunny, cloudy, light snow, etc.), thereby enhancing the transferability of the trained drone agent to real-world applications. Within the simulator, the drone can continuously navigate the urban environment we have loaded, capturing data visually through an integrated RGB, depth, and object segmentation cameras, which output corresponding first-person perspective images in real time.

\textbf{Scenes}. To bridge the gap between transferring drone intelligent agents from simulated environments to real-world scenarios, we utilize UrbanScene3D\cite{Liu2021UrbanScene3DAL}, a large-scale data platform specifically designed for urban scene perception and reconstruction. This platform comprises over 128,000 high-resolution images captured from various cities. The selected 3D scenes from four cities, as illustrated in Figure \ref{FIG:simulator}, feature detailed architectural elements, including office buildings, shopping centers, residential complexes, bus stations, and subway entrances and exits. Additionally, these scenes encompass specific street details such as lanes, sidewalks, crossroads, traffic signals, and road markings, along with other urban features like streetlights, signs, trees, shrubs, and lawns. These attributes facilitate the exploration of diverse urban environments by drone intelligent agents. Among the cities, Shanghai presents the most extensive urban scene, featuring $6,850$ objects and covering an area of $3,700$ hectares. This extensive environment is advantageous for training UAV agents in long-distance navigation and path planning. In contrast, the urban scene in Shenzhen is relatively compact, covering an area of $300$ hectares with only $1,126$ objects; however, it enhances the spatial reasoning capabilities of drone intelligent agents in smaller settings. Furthermore, the campus area, which spans $130$ hectares and contains $178$ objects, and the residential zone, covering $30$ hectares with $34$ objects, focus on localized environments characterized by dense buildings and obstacles such as trees and equipment. This concentration improves scene understanding and decision-making skills, including obstacle avoidance.

\textbf{Observations}. In the simulator, the drone is generated using AirSim, which features five built-in cameras: forward, backward, left, right, and overhead views. Each camera operates in three modes:

\textit{RGB Camera}. Captures RGB images with a resolution of $1920$$\times$$1080$, saved in PNG format.

\textit{Depth Camera}. Produces depth images based on the positional information between the camera and the object, maintaining the same resolution as the RGB camera and also saved in PNG format. In this experiment, when the distance exceeds 500 meters, the image appears entirely white; for distances below 500 meters, varying shades of black are displayed according to proximity.

\textit{Object Segmentation Camera}. Retrieves the object segmentation map, segmenting the image into different colors based on object types—gray for buildings, green for trees, and red for vehicles. The resolution of the segmentation image matches that of the RGB camera and is saved in PNG format.

\textbf{Actions}. The simulator supports drone intelligent agents in altering their position (x, y, z coordinates), direction (pitch, yaw, roll), and speed, while also enabling more complex maneuvers through acceleration adjustments and the application of force vectors. To facilitate the training of UAV agents, we have preliminarily identified the eight most common low-level actions for drones: forward, left turn, right turn, ascend, descend, left shift, right shift, and stop. To balance the frequency of actions during the trajectory with the actual movement of the drone in an outdoor environment, the “forward movement” action propels the drone continuously for $5$ meters in the current direction, while the “left movement” and “right movement” actions shift the drone continuously for $1$ meter in their respective directions. The left and right rotation actions enable horizontal rotation by $15$ degrees, and the ascending and descending actions allow vertical movement for $1$ meter.

\section{Dataset Suite}
\renewcommand{\dblfloatpagefraction}{0.9}
\begin{figure*}[t]
	\centering
		\includegraphics[scale=.23]{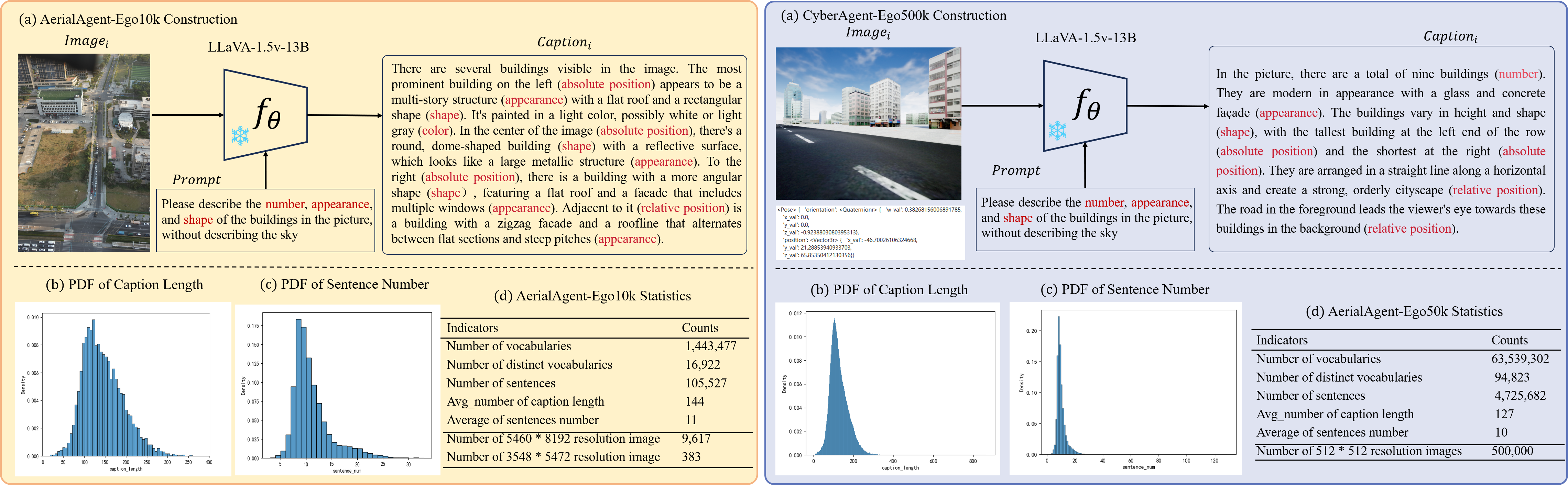}
	\caption{The left and right panels illustrate the construction schemes and statistics of the AerialAgent-Ego15k and CyberAgent-Ego500k datasets, respectively.
 }
	\label{FIG:AerialAgent-Ego10k_and_CyberAgent-Ego500k}
\end{figure*}
To address the shortage of large-scale training data for UAV agents, facilitate the training of aerospace embodied word models, and further advance research in aerospace embodied intelligence, we engage ten trained experts who dedicated eight months to developing a comprehensive dataset suite that encompasses two pre-training datasets and five downstream task instruction fine-tuning datasets.

\subsection{AerialAgent-Ego15k}
\textbf{Multi-Resolution UAV First-Person View City Images}. The first-person view images of real cities captured by drones are derived from the UrbanBIS dataset, which is collected using aerial photogrammetry and encompasses a wide array of urban scenes. Specifically, the UrbanBIS dataset\cite{Yang2023UrbanBISAL} comprises $0.5$ TB of aerial photographs from six actual locations: Qingdao, Wuhu, Longhua, Yuehai, Lihu, and Yingrenshi, covering a significant urban area of $10.78$ km² and including $3,370$ buildings, with a total of $113,346$ aerial photogrammetry images. We have requested images from the authors for the regions of Lihu, Longhua, Yingrenshi, and Yuehai, with resolutions of $6000 \times 4000$, $8192\times5640$, $5472\times3648$, and $5472\times3648$, respectively, yielding a total of $15,094$ images. From this dataset, we randomly selected $10,000$ images to serve as first-person view representations of real cities captured by drones.

\textbf{Fine-grained Multi-attribute First-view Text Generation}. To generate high-quality environmental descriptions, we utilize LLaVA-1.5-13B\cite{Liu2023ImprovedBW} to produce detailed accounts of surrounding buildings, roads, trees, and other scenery from first-person perspective images captured by a drone, as illustrated in Figure \ref{FIG:AerialAgent-Ego10k_and_CyberAgent-Ego500k} left (a). To standardize the format of the environmental descriptions generated by LLaVA-1.5-13B\cite{Liu2023ImprovedBW}, we employ specific prompts that emphasize the quantity, appearance, and shape of the buildings in the images, particularly focusing on the spatial relationships among the objects. This approach enhances the spatial reasoning capabilities of the drone agent. Furthermore, we specify that the sky should not be described, as this scene is relatively uniform and appears consistent from various perspectives of the drone, providing insufficient information. Consequently, the generated descriptions ensure a degree of diversity, accuracy, and detail.

\textbf{Diverse Data Distribution}. We perform a quantitative statistical analysis on AerialAgent-Ego15k. Figure \ref{FIG:AerialAgent-Ego10k_and_CyberAgent-Ego500k} (b) and (c) illustrate the probability density functions (PDFs) of text vocabulary length and text sentence length, respectively, both exhibiting a shape akin to a normal distribution. This finding supports the rationality of the text distribution. The maximum length for image descriptions is $440$ words, with an average length of $144$ words. The maximum number of sentences in image descriptions is $42$, with an average of $11$ sentences per image. Both the number of sentences and text lengths exceed those of most existing visual-language datasets. Figure \ref{FIG:AerialAgent-Ego10k_and_CyberAgent-Ego500k} (d) reveals that the dataset contains a total of $158,379$ sentences and $2,167,455$ words, of which $21,489$ are unique.

\subsection{CyberAgent-Ego500k}
\textbf{Image Acquisition}. We require trained drone pilots to operate drones in four virtual cityscapes: Shenzhen, School, Residence, and Shanghai. The flight range encompasses the entirety of these city scenes, with dense sampling conducted in areas characterized by a high density of objects, such as buildings. To prevent the drones from encountering obstacles, a selection of drone poses is recorded at random. Based on these poses, a total of $1,040,924$ first-person perspective images, each with a resolution of $512 \times 512$ pixels, are generated within the virtual cityscapes. From this collection, $500,000$ images are randomly selected to construct the image-text-pose dataset.

\textbf{First-Person Image-Text-Pose Generation}. As illustrated on the right side of Figure \ref{FIG:AerialAgent-Ego10k_and_CyberAgent-Ego500k} (a), the dataset construction method aligns with that of AerialAgent-15k and exhibits the following three characteristics:

\textit{Drone First-Person Images in Multi-City Scenes}. Collected from 3D simulators in Shanghai (large areas), Shenzhen (multiple blocks), campuses (featuring numerous obstacles such as trees), and residential areas (characterized by dense buildings and narrow pathways), this approach aims to minimize the gap between simulated and real-world environments.

\textit{Multi-Attribute First-Person Text Descriptions}. The generated text descriptions provide comprehensive information regarding the attributes of objects in the drone’s first-person images, including appearance, quantity, shape, absolute position, and relative position. Notably, the spatial relationships among objects are crucial for enhancing the spatial reasoning capabilities of the drone agent.

\textit{Image-Text-Pose Alignment}. In addition to the images and their corresponding text descriptions, this method incorporates the drone’s attitude (position and orientation) in 3D space. The objective is to integrate the drone’s spatial positioning into the aerospace-embodied world model, thereby enhancing the drone’s self-centered scene understanding capabilities.

\textbf{Dataset statistics}. Figure \ref{FIG:AerialAgent-Ego10k_and_CyberAgent-Ego500k} (b), (c), and (d) in the right block present detailed statistical results for the CyberAgent-500k dataset. The maximum length of the image descriptions is $865$ words, with an average length of $127$ words. Furthermore, the maximum number of sentences per image description is $129$, with an average of $10$ sentences. The dataset contains a total of $4,725,682$ sentences and $63,539,302$ words, including $94,823$ unique words. These statistical results indicate that this dataset surpasses most existing visual-language datasets in terms of scale, text length, sentence count, and the alignment of drone poses.

\subsection{SkyAgent-Scene3k}
\renewcommand{\dblfloatpagefraction}{0.9}
\begin{figure*}[t]
	\centering
		\includegraphics[scale=.3]{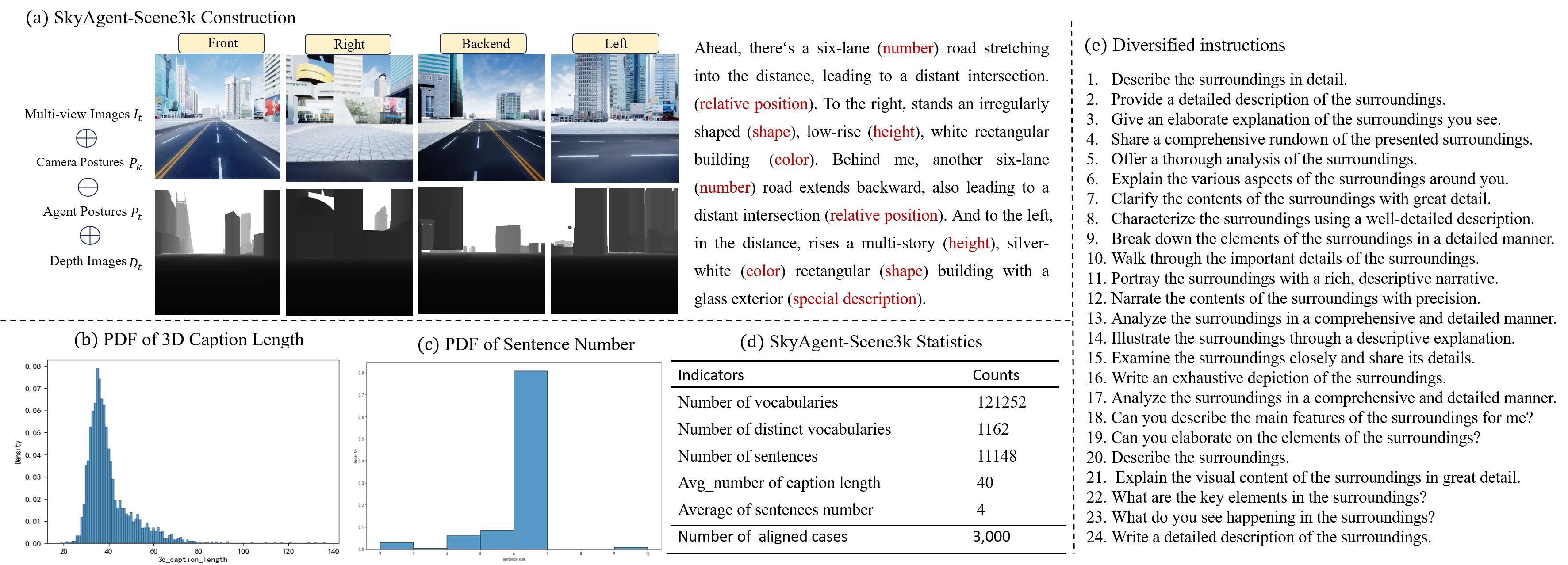}
	\caption{SkyAgent-Scene3k dataset of concrete examples, statistical results, and diversified instructions.
 }
	\label{FIG:SkyAgent-Scene3k}
\end{figure*}

\textbf{Dataset Construction}. We require the annotator to control the drone to navigate within the 3D virtual city scene, select its current posture, and describe the surrounding environment from four perspectives: front, back, left, and right. The description format is fixed as follows: “front $\langle$ object description $\rangle$, right $\langle$ object  description $\rangle$, back $\langle$  object  description $\rangle$, left $\langle$ object  description $\rangle$”. The object description should include the elements “quantifier + color + specific description + shape + object”, as illustrated in Figure \ref{FIG:SkyAgent-Scene3k} (a). To ensure data quality, we conduct rigorous inspections, requiring different annotators performing the same task to cross-check their work, followed by cross-checking between annotators from different cities. In summary, SkyAgent-Scene3k possesses the following characteristics:

\textit{Diversified Object Types and Instructions}. The primary objects include buildings, roads, trees, and grasslands within urban areas. Additionally, we have developed over 20 distinct instructions, as illustrated in Figure \ref{FIG:SkyAgent-Scene3k} (e), to enhance the generalization capabilities of task understanding.

\textit{Multi-Directional and Multi-Attribute Environment Description}. Focusing on the drone intelligent agent, descriptions of both close-range and long-range scenes are provided from four perspectives: front, back, left, and right. Buildings are characterized by their height, appearance, and color, while roads are described based on the number of lanes, intersections, and directional extensions.

\textit{Multi-Perspective 2D Images, Depth Maps, Camera Poses, Drone Poses, and Scene Description Alignment}. Multi-perspective images, depth maps, and camera poses of urban landscapes facilitate the reconstruction of a three-dimensional representation of the entire city, assisting drone agents in understanding the spatial relationships between objects and enhancing their perception of three-dimensional scenes.

\textbf{Dataset statistics}. Figure \ref{FIG:SkyAgent-Scene3k} (b), (c), and (d) illustrate the distribution of description lengths, the number of sentences, and statistical information regarding scene descriptions. As shown in Figure \ref{FIG:SkyAgent-Scene3k} (b), the lengths of the descriptions range from $30$ to $80$ words. Generally, longer descriptions suggest a more complex scene with a greater number of environmental elements that require articulation. Figure \ref{FIG:SkyAgent-Scene3k} (c) indicates that most descriptions consist of four sentences, as we instruct annotators to depict each scene from four perspectives: front, back, left, and right. Descriptions containing 1$\sim$3 sentences occur when annotators consolidate multiple perspectives into a single sentence. In total, this dataset comprises $121,252$ words and $1,162$ distinct word types.

\subsection{SkyAgent-Reason3k}
\renewcommand{\dblfloatpagefraction}{0.9}
\begin{figure*}[t]
	\centering
		\includegraphics[scale=.3]{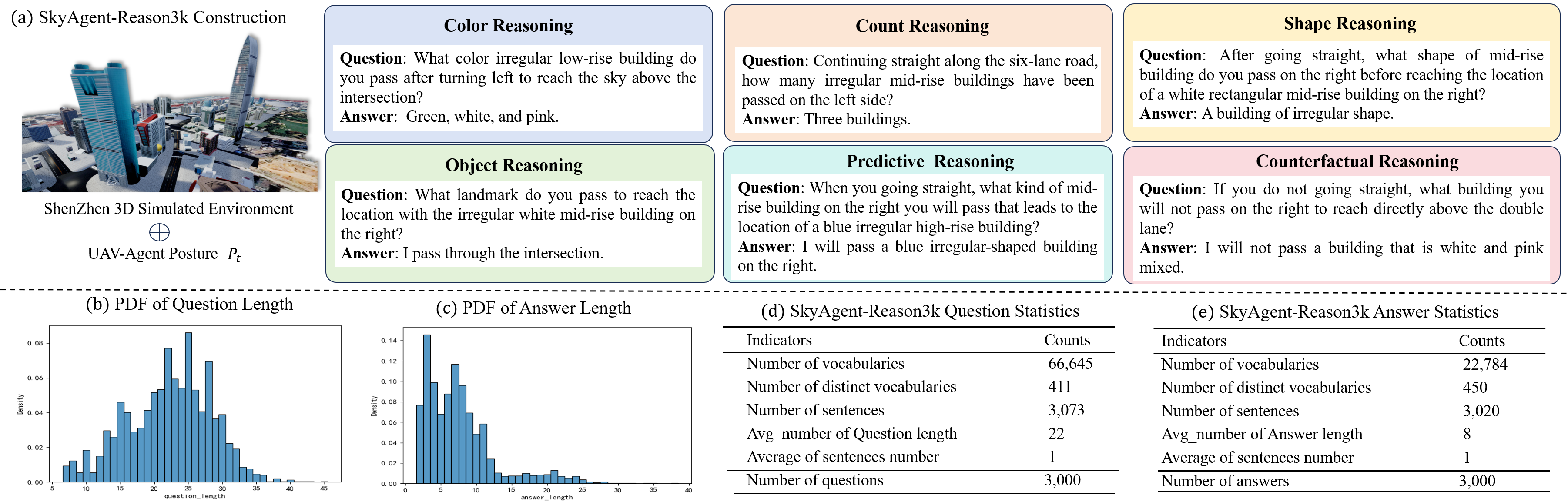}
	\caption{The SkyAgent-Reason3k dataset contains $6$ types of reasoning patterns and corresponding concrete examples, as well as some statistical results. 
 }
	\label{FIG:SkyAgent-Reason3k}
\end{figure*}
\textbf{Dataset Construction}. To enhance the cognitive reasoning abilities of UAV agents in three-dimensional urban environments, we require annotators to navigate the 3D city scene, adopt specific postures to pause, establish targeted spatial positions, and create question-and-answer pairs regarding various features encountered by the UAV, including buildings, roads, trees, and grasslands. Specifically, inquiries pertaining to buildings should focus on attributes such as height, appearance, and color, while questions related to roads should address the number of lanes, intersections, and direction of extension. As illustrated in Figure \ref{FIG:SkyAgent-Reason3k} (a), each question in this dataset must be answered accurately through spatial reasoning in conjunction with the three-dimensional environment. This process can be further categorized into six distinct modes of reasoning.

\begin{itemize}
    \item \textit{Color Reasoning}. This reasoning process involves prompting the drone’s intelligent agent to identify and inquire about the colors of specific objects encountered as it approaches a designated spatial location. This necessitates the agent’s ability to recognize colors based on the identified targets.
    \item \textit{Count Reasoning}. Requires the intelligent agent to compute the number of specific objects encountered while following short-range instructions.
    \item \textit{Shape Reasoning}. This reasoning necessitates that the drone’s intelligent agent describes the specific shapes of the objects it encounters upon arriving at the designated target area.
    \item \textit{Object Reasoning}. Requires UAV intelligent agents to enumerate the buildings and other objects they encounter while navigating to a specific spatial location.
     \item \textit{Predictive Reasoning}. Upon satisfying certain preconditions, the drone must predict potential objects and actions it may encounter.
     \item \textit{Counterfactual Reasoning}: This reasoning involves presenting a hypothesis to the drone agent that contradicts established facts, requiring the agent to respond to the hypothesis.
\end{itemize}

\textbf{Dataset Statistics}. Figure \ref{FIG:SkyAgent-Reason3k} (b) illustrates that the length distribution of questions ranges from $7$ to $45$ words, significantly surpassing the statistics of questions found in existing VQA datasets in terms of both coverage and length. Figure \ref{FIG:SkyAgent-Reason3k} (c) indicates that the length of answers varies from $2$ to $40$ words, with the majority consisting of $2$ to $10$ words, thereby allowing the drone agent to deliver concise responses. Figures \ref{FIG:SkyAgent-Reason3k} (d) and (e) present statistical analyses of the questions and answers, respectively. The results reveal that, although the word count of the questions is approximately three times greater than that of the answers ($66,645$ vs. $22,784$), the vocabulary diversity is actually lower in the questions than in the answers ($411$ vs. $450$). This discrepancy underscores the potential for drone intelligent agents to enhance their vocabulary in responses.

\subsection{SkyAgent-Nav3k}
\renewcommand{\dblfloatpagefraction}{0.9}
\begin{figure*}[t]
	\centering
		\includegraphics[scale=.28]{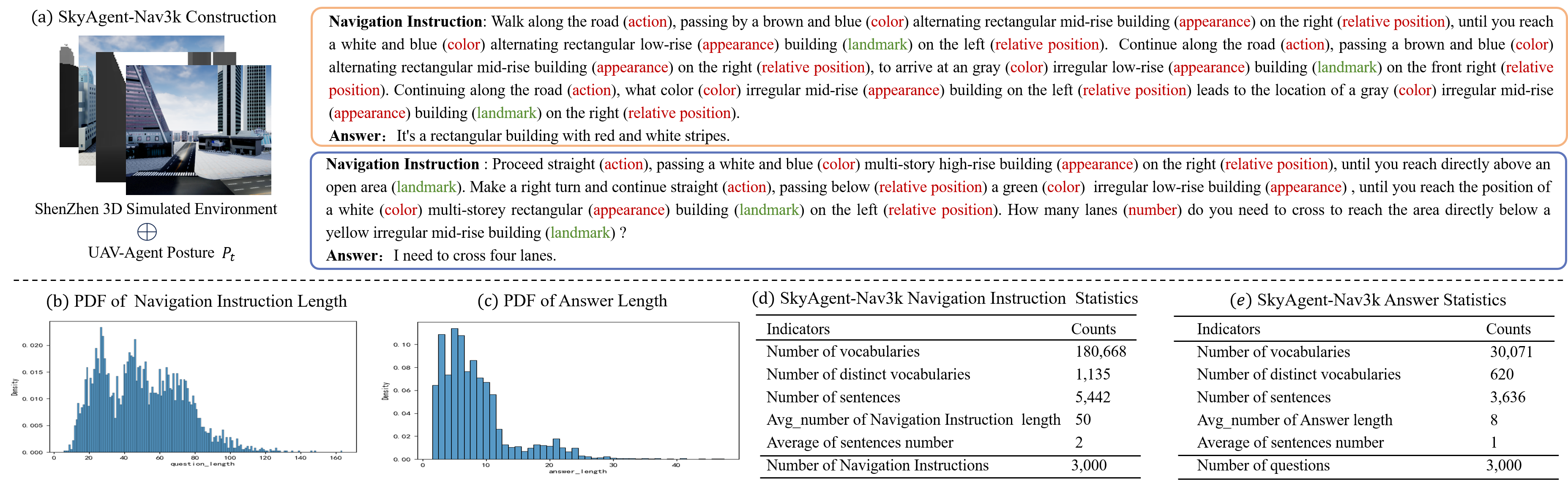}
	\caption{Two concrete examples selected from the SkyAgent-Nav3k dataset and the statistical results. 
 }
	\label{FIG:SkyAgent-Nav3k}
\end{figure*}
\textbf{Dataset Construction}. We require annotators to control drones to fly specific distances within an urban environment, annotate the textual descriptions of the flight paths, record the starting and ending positions, and design a set of question-and-answer pairs primarily addressing whether actions such as flying straight, turning left, or turning right will occur, as well as the types of buildings, intersections, and lanes encountered. Additionally, manual cross-validation is employed to ensure the quality of the annotations. Two specific examples are illustrated in Figure \ref{FIG:SkyAgent-Nav3k} (a), from which the following characteristics of the dataset can be derived:

\textit{Refined Object Attribute Description and Navigation Instructions}. The navigation instructions provide comprehensive descriptions of the object, detailing its appearance, quantity, shape, color, and relative position to the drone’s intelligent agent. This ensures the uniqueness of the object in the instructions and minimizes the error recognition rate.

\textit{Long-Range Navigation Path Guided by Multiple Landmarks}. The navigation instructions encompass extended paths that necessitate multiple consecutive spatial inferences by drones to traverse various blocks within the city. Furthermore,  the instructions include specific descriptions of landmarks that can assist the drone’s intelligent agent in adjusting its actions.

\textit{Navigation-Based Scene Exploration}. In addition to requiring the drone to adhere to language instructions for navigating to a designated location, this dataset also compels the drone agent to articulate environmental information regarding the destination, such as the color and shape of buildings.

\textbf{Dataset Statistics}. From Figure \ref{FIG:SkyAgent-Nav3k} (b), it is evident that the length of navigation instructions predominantly ranges from $20$ to 80, exhibiting a relatively even distribution, with a few instances exceeding $100$, which surpasses most existing navigation datasets. Longer navigation instructions can enhance drones’ long-range spatial reasoning abilities. Figure \ref{FIG:SkyAgent-Nav3k} (c) indicates that the lengths of answers primarily fall between $2$ and $10$, facilitating drone agents in succinctly describing objects to be explored. Figures \ref{FIG:SkyAgent-Nav3k} (d) and (e) present statistical analyses of the navigation instructions and answers, revealing average lengths of $50$ and $8$, respectively, with an average of $2$ sentences for navigation instructions and 1 sentence for answers. This variance arises because navigation commands encompass both long-distance, multi-step instructions and attribute queries regarding unknown objects.

\subsection{SkyAgent-Plan3k}
\renewcommand{\dblfloatpagefraction}{0.9}
\begin{figure*}[t]
	\centering
		\includegraphics[scale=.27]{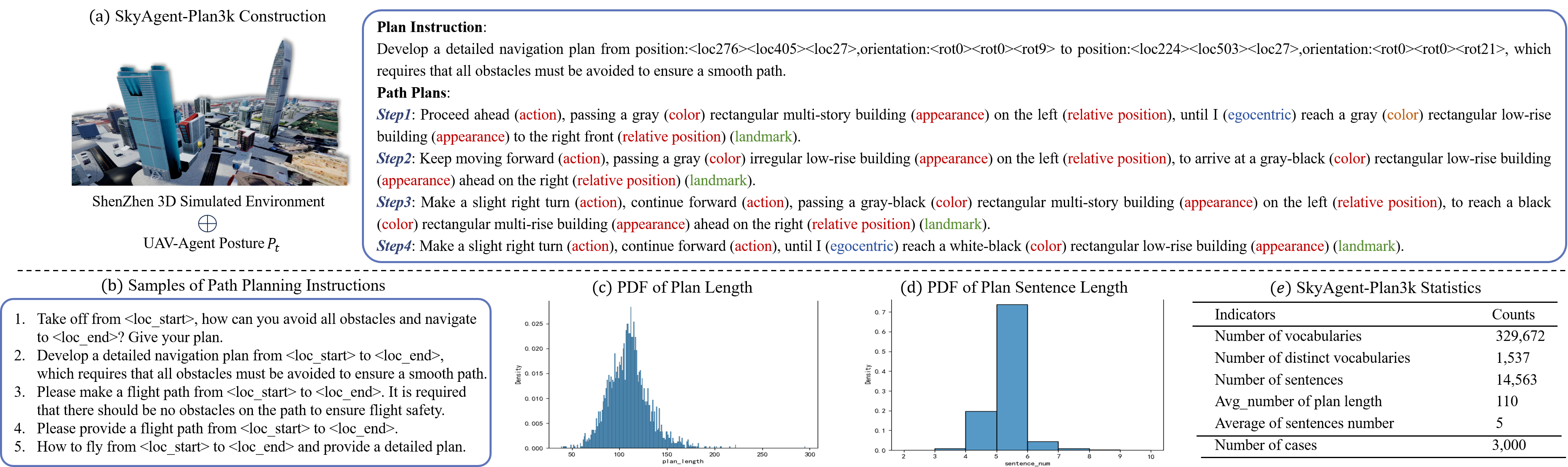}
	\caption{A example selected from the SkyAgent-Plan3k dataset and the statistical results, as well as several path planning instructions. 
}
	\label{FIG:SkyAgent-Plan3k}
\end{figure*}
\textbf{Dataset Construction}. We require drone pilots to identify the starting and ending points prior to operating the drone. After flying for a specified duration, they should select a position that serves as the midpoint of the trajectory and provide a description of the route from the previous trajectory to the current location. To generate high-quality route descriptions, we ask drone pilots to choose the optimal path based on their experience. Furthermore, we require professional annotators to provide detailed descriptions of sub-routes in specific scenarios, such as making turns, navigating intersections, or passing by five buildings in a single direction. Figure \ref{FIG:SkyAgent-Plan3k} (a) illustrates an example of path planning, demonstrating the following characteristics:

\textit{Refined Self-Centered Object Description}. The drone agent provides a distinctive and identifiable description of objects based on color, shape, height, and structure, employing a first-person perspective. The objects include buildings, pathways, trees, and grasslands that sequentially appear on both the left and right sides.

\textit{Multi-Perspective Object Localization}. In three-dimensional urban environments, the UAV agent accurately locates instance-level objects, such as buildings, by establishing spatial relationships relative to itself, thereby enhancing the precision of object localization.

\textit{Landmark-Guided Path Planning}. Prior to executing maneuvers such as turning or proceeding straight, the UAV intelligent agent identifies a landmark as a reference point, thereby improving the accuracy of path planning.

\textbf{Dataset Statistics}. Figure \ref{FIG:SkyAgent-Plan3k} (b) presents several instructions for the drone agent concerning path planning, each requiring the agent to avoid obstacles while navigating from the starting point to the endpoint. Figure \ref{FIG:SkyAgent-Plan3k} (c) illustrates that the planned lengths range significantly from $25$ to $225 $ and predominantly follow a normal distribution. Figure \ref{FIG:SkyAgent-Plan3k} (d) indicates that the majority of the dataset consists of planning for five sub-paths. This requirement is designed to enhance planning complexity, necessitating the drone to perform at least five actions and navigate over five objects, thereby improving its capability to plan for longer distances. Figure \ref{FIG:SkyAgent-Plan3k} (e) reveals that the average length of the plans is $110$, which is generally higher than the task planning lengths observed in most indoor scenarios.

\subsection{SkyAgent-Act3k}
\renewcommand{\dblfloatpagefraction}{0.9}
\begin{figure*}[t]
 	\centering
 		\includegraphics[scale=.31]{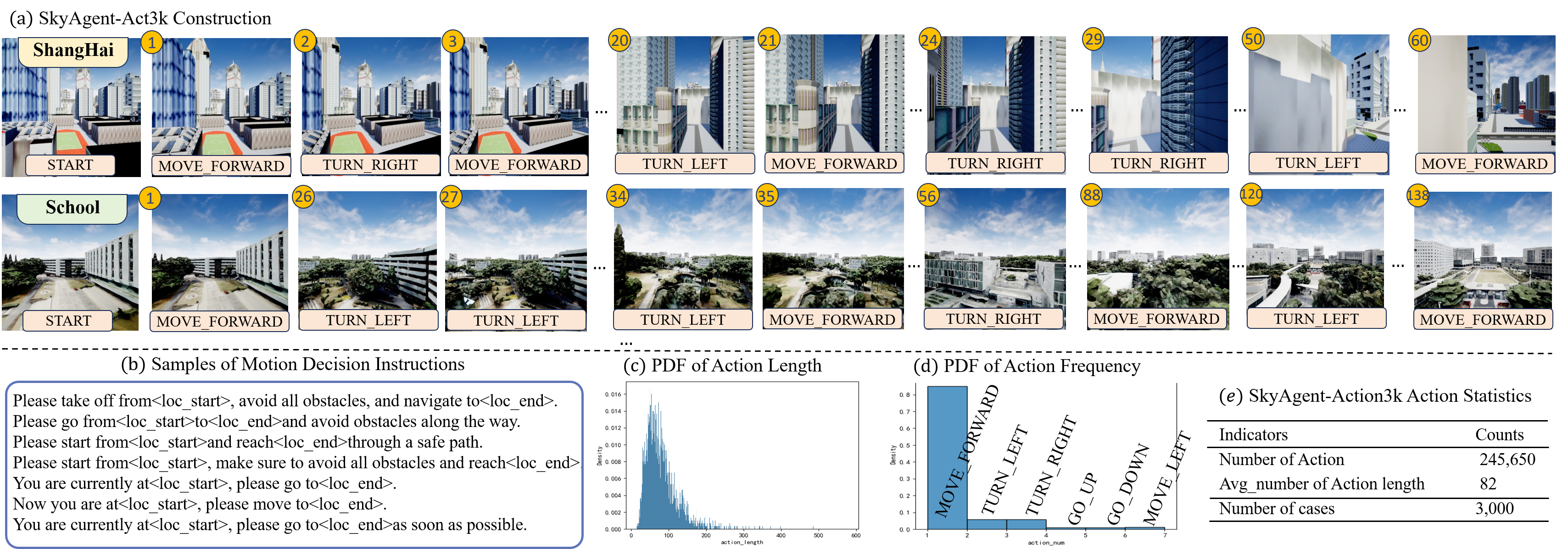}
 	\caption{Examples of drone agent actions in $2$ city scenarios (i.e., ShangHai and School) in SkyAgent-Act3k dataset, along with some instructions and statistics. 
  }
 	\label{FIG:SkyAgent-Act3k}
 \end{figure*}
\textbf{Dataset Construction}. This task involves recording the dense motion sequence and orientation of the drone, with a particular emphasis on its flight path. Consequently, we restrict the drone’s flight altitude to within 30 meters. The drone pilot is required to select both the starting and ending points, maneuver the drone to depart from the starting location, and leverage their experience to choose an appropriate route to reach the destination. This process allows us to capture the starting point, ending point, drone orientation, and action sequence. To ensure a high-quality path of reasonable length, we instruct drone pilots to avoid choosing arbitrary routes, such as unnecessary detours. Additionally, the need for drone pilots to survey their surroundings to ascertain their position and determine the next destination may lead to excess motion. We mitigate this excess motion through post-processing to achieve a smoother trajectory. Figure \ref{FIG:SkyAgent-Act3k} (a) illustrates a series of drone action decisions, which exhibit the following characteristics:

\textit{Starting and Ending Points Beyond Visual Range}: To enhance the long-range autonomous action control capability of UAV intelligent agent in large-scale urban environments, there must be a minimum of ten buildings situated between the starting and ending points, with these buildings not aligned on the same straight line. This necessitates that the UAV intelligent agent execute at least one turn.

\textit{Professional Path Selection}: Upon determining the starting and ending points, the drone pilot selects the optimal flight route based on experience, while ensuring that the flight altitude does not exceed 30 meters. The route selection must avoid collisions with surrounding objects and unnecessary turns and detours.

\textit{Smooth Action Sequence}: The drone pilot consciously avoids sharp turns, emergency stops, and abrupt maneuvers when performing turns, ascents, and other actions during flight, striving to ensure smooth transitions in the drone’s movements.

\textbf{Dataset Statistics}. Figure \ref{FIG:SkyAgent-Act3k} (b) presents several examples of motion decision-making instructions, illustrating that these instructions primarily convey the requirement for the drone to navigate obstacles safely, quickly, and autonomously from the starting point to the endpoint. Figure \ref{FIG:SkyAgent-Act3k} (c) indicates that the lengths of motion sequences in the dataset predominantly range from 50 to 150, significantly exceeding the action lengths of intelligent agents in existing indoor scenarios. Figure \ref{FIG:SkyAgent-Act3k} (d) illustrates the distribution of various actions, revealing that “MOVE-FORWARD” is considerably more prevalent than other actions. This observation is entirely logical, as the process of flying from the starting point to the endpoint involves primarily forward flight, with turns, ascents, and other maneuvers required to avoid obstacles and detours. The average length of the action sequences depicted in Figure \ref{FIG:SkyAgent-Act3k} (e) is $82$, further emphasizing that our dataset focuses on long-distance drone flights in large-scale urban environments. This distinction marks the primary difference between this dataset and those associated with existing indoor scene datasets.
\subsection{Image Enhancement}
\begin{figure}[t]
 	\centering
 		\includegraphics[scale=.6]{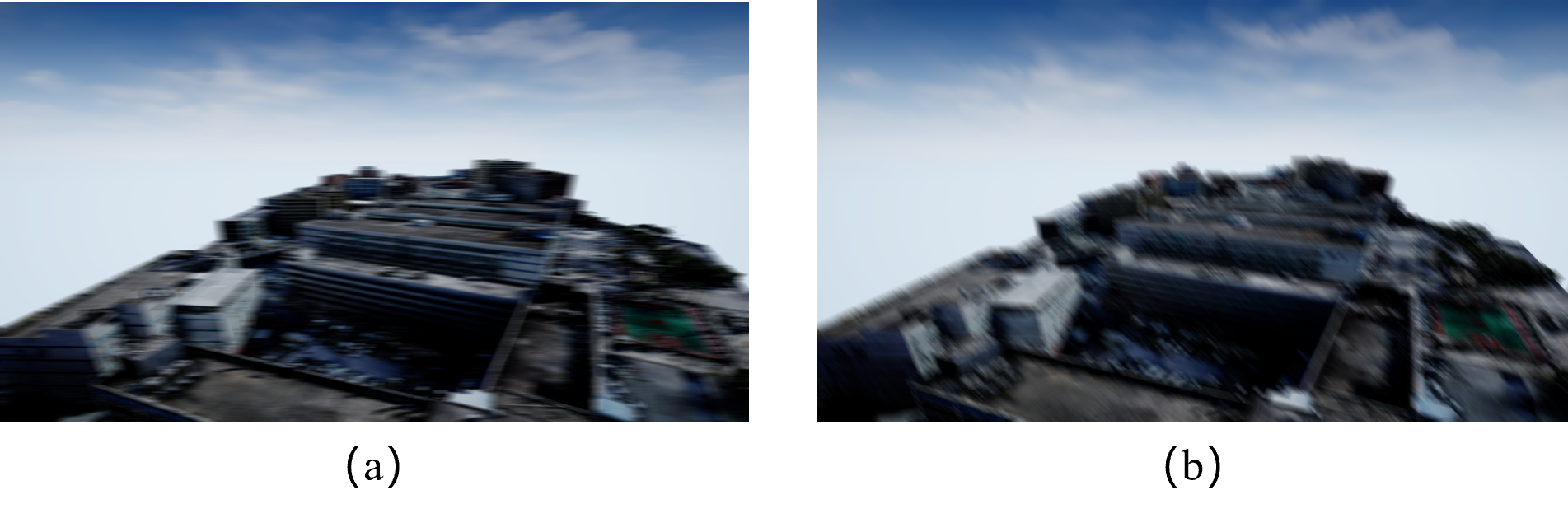}
 	\caption{Figure (a) and (b) show the motion blur effects in the horizontal direction and the 45-degree direction, respectively.
  }
 	\label{FIG:Enhancement}
 \end{figure}

To fully consider the challenges faced by drones in real-world complex environments, this study systematically addressed two key issues encountered during flight through data augmentation. First, regarding the image tilt caused by drone body jitter, we collect more diverse data on camera tilts in the environment. Second, to address image blurring or pose drift due to jitter, we simulate the motion blur degradation process of drone images by constructing directional motion blur kernels (PSF) and applying convolutional operations. Based on the assumption of linear uniform motion, the algorithm first generates blur kernels of specified size and angle, ensuring energy conservation through normalization, and then employs two-dimensional discrete convolution to achieve image degradation. This method effectively simulates typical motion blur caused by drone body jitter or pose drift, as shown in figure \ref{FIG:Enhancement}.
\renewcommand{\dblfloatpagefraction}{0.9}

\section{Experiments}
\subsection{Baselines}
\textbf{Baselines Selection}. Due to the current scarcity of research on aerospace-embodied world models, we evaluate several mainstream and representative 3D and 2D visual-language models. This assessment aims to explore their potential and limitations concerning the proposed aerospace-embodied downstream task datasets, thereby providing a preliminary foundation for future researchers in the field of aerospace-embodied intelligence. While there are more 2D visual-language models available that are generally more mature, we focus on LLaVA\cite{Liu2023VisualIT}, MiniGPT4\cite{Zhu2023MiniGPT4EV}, and BLIP2\cite{Li2023BLIP2BL}, categorizing them into $7B$ and $13B$ models based on parameter scales. Given the limited availability of open-source 3D visual-language models, we select only the 3D-LLM\cite{Hong20233DLLMIT} as our research focus. %Specific information is provided in Table \ref{TAB:Baselines}.

\renewcommand{\dblfloatpagefraction}{0.9}
\begin{figure}[t]
	\centering
		\includegraphics[scale=.2]{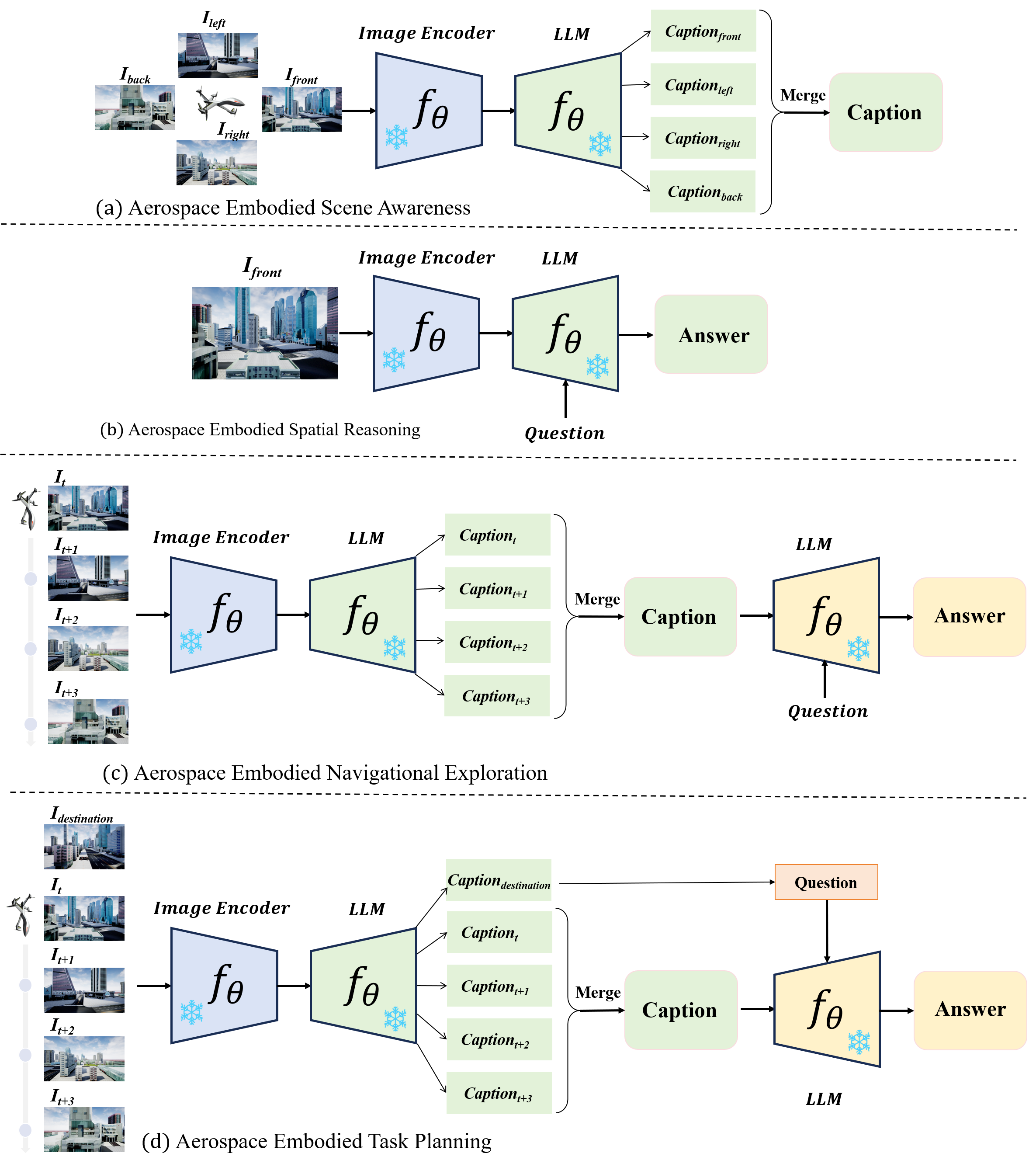}
	\caption{Specific modifications of visual-language models for aerospace embodied downstream tasks . 
 }
	\label{FIG:task-modification}
\end{figure}

\textbf{Baselines Modification}\label{Baselines_Modification}. Among the selected baseline models, the 3D visual-language model can be applied to most of the defined downstream tasks; however, the 2D visual-language model cannot be directly utilized for testing due to a mismatch in input formats, as illustrated in Figure \ref{FIG:task-modification}. Consequently, we modify the inputs and outputs of these models to align with the downstream tasks, as detailed below. Notably, Aerospace Embodied Motion Decision represents the culmination of aerospace embodied tasks, achieving a closed loop of perception, cognition, and action for the UAV agent. Adjusting existing visual-language models presents challenges, and we will continue to explore this area in future.

\textit{Aerospace Embodied Scene Awareness}. This task involves utilizing the location and environmental data captured by the drone as input to generate scene descriptions of the surrounding environment from multiple perspectives. However, the 2D visual-language model is inherently limited to processing images and does not directly account for environmental features. To mitigate this limitation during testing, we modify the 2D visual-language model by providing it with four images captured from the drone’s perspectives: front, back, left, and right. After generating captions for these images using descriptive prompts, we concatenate the four captions to produce the output for environmental observation, as illustrated in Figure \ref{FIG:task-modification} (a).

\textit{Aerospace Embodied Spatial Reasoning}. This task also requires the integration of 3D features; thus, we modify the 2D visual-language model during testing by adjusting the input to include both the observation image and the question presented directly in front of the drone’s position. By reasoning and responding to questions based on this image, we generate spatial reasoning answers, as illustrated in Figure \ref{FIG:task-modification} (b).

\textit{Aerospace Embodied Navigational Exploration}. As illustrated in Figure \ref{FIG:task-modification} (c), the input consists of multiple images and questions along the drone’s flight path. After generating captions for each image, the questions are answered based on the concatenated captions, ultimately yielding the solution for the drone’s navigation exploration.

\textit{Aerospace Embodied Task Planning}. As illustrated in Figure \ref{FIG:task-modification} (d), we modify the input to encompass multiple images depicting the drone’s flight path, in addition to the endpoint image. Initially, a caption for the endpoint image will be generated, followed by the formulation of a question directed at the drone’s intelligent agent, inquiring about the navigation method to reach the specified location. Subsequently, the answer for the drone’s path planning will be derived based on the caption of the concatenated flight path images.

\subsection{Evaluation Metrics}
\textbf{Traditional Metrics}. Common indicators include \textit{BLEU-1}, \textit{BLEU-2}, \textit{BLEU-3}, and \textit{BLEU-4}\cite{2002BLEU}. Compare the degree of overlap between the n-grams in the candidate translation and the reference translation. It is commonly used for evaluating translation quality and can be divided into multiple evaluation indicators based on n-grams. 

\textit{CIDEr}\cite{Vedantam2014CIDErCI} is an evaluation metric used to assess image description tasks. Its main idea is to treat each sentence as a document, then calculate its n-gram TF-IDF vector, and use cosine similarity to measure the semantic consistency between candidate sentences and reference sentences.

\textit{SPICE} \cite{Anderson2016SPICESP} utilizes graph-based semantic representations to encode objects, attributes, and relationships within descriptions. Initially, it parses both the description under evaluation and the reference description into a syntactic dependency tree using a Probabilistic Context-Free Grammar (PCFG) dependency parser. %Subsequently, it maps the dependency tree into a scene graph via a rule-based approach. Finally, it calculates the F-score values for the objects, attributes, and relationships in the description being evaluated.

% \textit{KAS} focuses on the key actions that generate the plan, such as "turn left", "turn around", etc., and checks whether they are part of the reference sentence to measure the accuracy of the key actions.

% \textit{Navigation Error} (NE) refers to the geodesic distance between an agent’s final position and the target.

% \textit{Success Rate} is defined as the percentage of stops that fall within the threshold distance of the target. A navigation is deemed successful if the drone’s stopping position accurately corresponds to the target position and the distance is within the specified threshold.

% \textit{Oracle Success Rate} pertains to the assessment of success for any point along the predicted trajectory that lies within a threshold distance from the target location.

% \textit{Success Rate Weighted by Normalized Dynamic Time Warping} (SDTW) incorporates both the navigation success rate and the similarity between the actual path taken and the path predicted by the model.

\textbf{GPT4-based Metrics}. GPT-4\cite{Achiam2023GPT4TR} has achieved significant success in aligning with human preferences. Consequently, we introduce an automated evaluation method based on GPT-4 for tasks related to aerospace embodied scene awareness, spatial reasoning, navigational exploration, and path planning. This method aims to produce evaluation results that closely resemble human assessments. By designing various prompt templates, we can effectively address different evaluation concerns.

\textit{LLM-Judge-Scene}. Aerospace Embodied Scene Awareness requires the intelligent drone agent to describe the scene from multiple perspectives. Therefore, the design of the evaluation method must consider both the level of detail in the descriptions and their relevance to the specified direction. To achieve this, we have developed a prompt template for GPT-4 that separately scores the granularity of the descriptions and the accuracy of each directional response.

\textit{LLM-Judge-Reason$\&$Nav}. The prompt language is aligned with that of llm-judge\cite{Zheng2023JudgingLW}, enabling GPT-4 to analyze the correlation and utility between AI assistant responses and reference answers. This process aims to objectively identify and correct errors to the greatest extent possible, provide explanations, and ultimately assign scores.

\textit{LLM-Judge-Plan}. Certain key actions in the plan, such as left and right turns, are critical, particularly concerning their sequence. Additionally, accurately describing the path requires noting significant buildings and landmarks along the route. To enhance the effectiveness of GPT-4 in scoring the generated responses, we have directed it to focus on two aspects: (a) the degree of alignment between the key action sequence and the reference answer, and (b) the accuracy of the descriptions of the buildings along the route, including their order and direction of passage.

\textbf{Human Evaluation}. While automated metrics provide scalable and consistent evaluation, they may not fully capture the nuances of human judgment, especially for complex cognitive tasks like spatial reasoning and path planning where logical coherence and real-world feasibility are paramount. To provide a more holistic assessment and to validate our automated findings, we complement our quantitative analysis with targeted human evaluation. Given the large scale of our benchmark, which makes comprehensive manual annotation infeasible, we conduct our evaluation on a randomly sampled subset of 100 instances for each of the four downstream tasks. We enlist experts to score the model-generated responses on a scale of 0 to 1 based on the following task-specific criteria:
For Scene Awareness, evaluators assess the accuracy and completeness of object descriptions, focusing on attributes like color, shape, and quantity.
For Spatial Reasoning, scoring is based on the logical correctness of the inferred relationships between objects.
For Navigational Exploration, judgment considers both the accuracy of following the navigation command and the correctness of the answer provided.
For Task Planning, scores are assigned based on the plan's feasibility, the correctness of the action sequence, and the accuracy of landmark identification.

\section{SkyAgentX: unifying perception, reasoning, planning,  and navigating}

\renewcommand{\dblfloatpagefraction}{0.9}
\begin{figure*}[t]
 	\centering
 		\includegraphics[scale=.65]{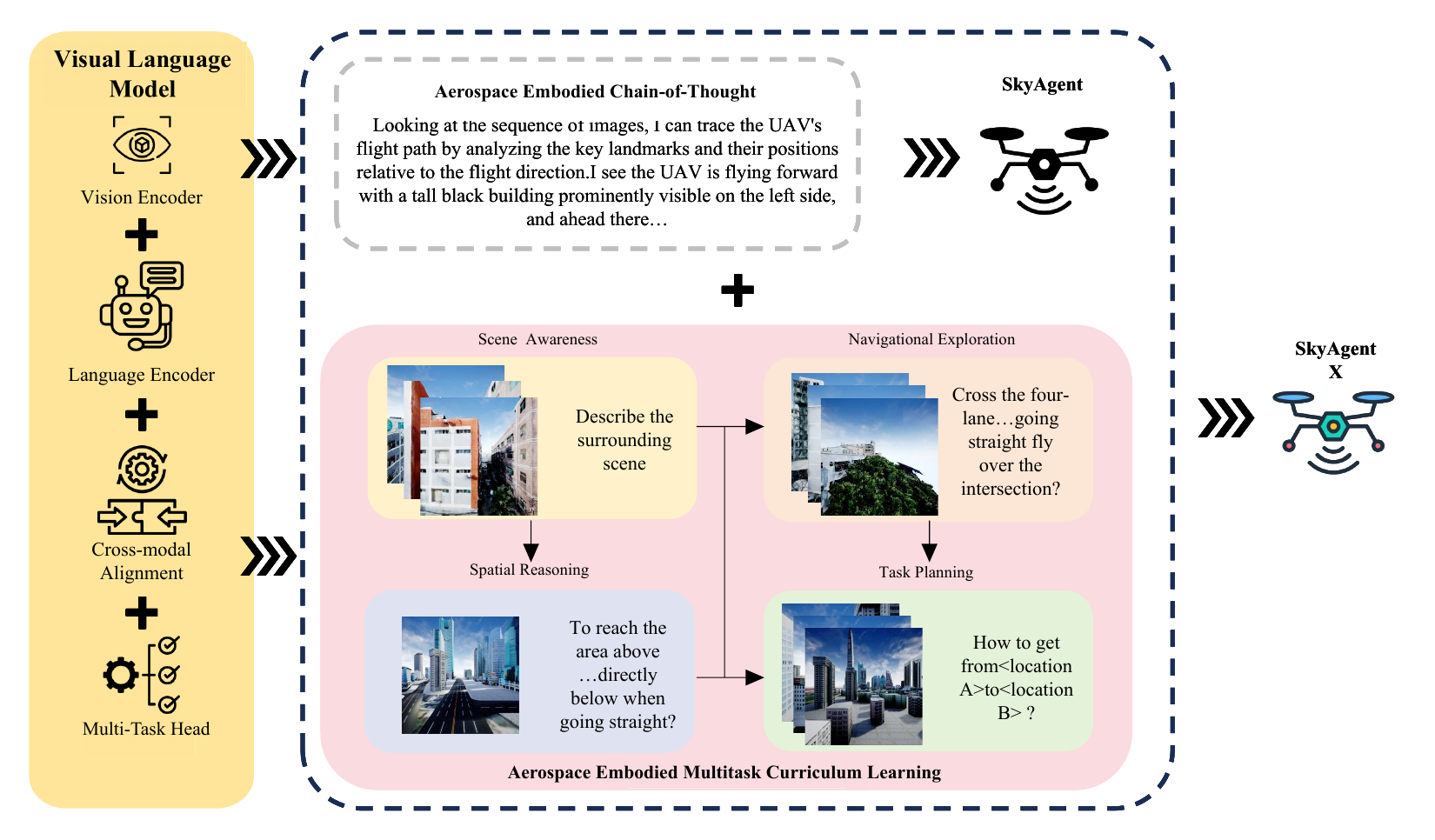}
 	\caption{Overview of the UAV-agent embodied large model, SkyAgentX, which integrates “Perception-Reasoning-Navigating-Planning” into an unified framework with aerospace embodied chain-of-thought and multitask curriculum learning.}
 	\label{FIG:model_framework}
 \end{figure*}

\subsection{Aerospace Embodied Chain-of-thought}

Chain-of-thought is a mechanism that mimics the human step-by-step reasoning process. By breaking down complex problems into multiple intermediate steps, it guides the model to generate coherent and reasonable answers. Chain-of-thought explicitly displays the model's reasoning path, making the decision-making process transparent and facilitating the analysis of error sources. For example, in drone mission planning, the model can first identify environmental features, then evaluate feasible paths, and finally generate action sequences, with each step being traceable, thereby enhancing the model's interpretability. By solving sub-problems step by step, the model can more efficiently handle multimodal, long-sequence, or highly abstract tasks.

In the AeroVerse benchmark, we automatically expand data through large models (such as GPT-4), generate diverse reasoning paths through prompt engineering, and construct aerospace embodied Chain-of-thought training data.

\subsection{Aerospace Embodied Multitask Curriculum Learning}
In the AeroVerse benchmark, the design of the five downstream tasks (aerospace scene perception, spatial reasoning, navigational exploration, task planning, and motion decision-making) follows a progressive relationship from simple to complex and from perception to decision-making. This structure naturally lends itself to the training strategy of Curriculum Learning, which involves phased, incremental task training to gradually enhance the comprehensive capabilities of the drone agent.

As the starting point of Curriculum Learning, Aerospace Embodied Scene Awareness requires the drone to describe its surroundings (e.g., buildings, roads, trees, etc.) from a first-person perspective. The goal of this stage is to equip the model with basic environmental understanding capabilities. This phase resembles the “observation and description” stage in human learning, providing the foundation of environmental cognition for subsequent tasks. Through extensive scene description training, the model establishes preliminary representational abilities for three-dimensional space. 

After mastering scene perception, the model must further understand the spatial relationships between objects. The Aerospace Embodied Spatial Reasoning task requires the drone to answer questions about the environment, upgrading from static descriptions to dynamic reasoning. The model must perform logical judgments by combining three-dimensional scene features (e.g., depth information, camera pose). This stage reinforces the model's spatial modeling and causal reasoning capabilities through complex questions in the SkyAgent-Reason3k dataset, laying the groundwork for subsequent navigation tasks. 
The Aerospace Embodied Navigational Exploration task requires the drone to explore the environment based on long-distance navigation instructions (e.g., “\textit{Fly forward 200 meters and then turn left}”) and answer attribute-related questions about objects encountered along the way. The navigation task introduces a temporal dimension and action sequences, requiring the model to translate the perception and reasoning abilities learned in the previous stages into concrete actions. Through training on the SkyAgent-Nav3k dataset, the model learns to achieve a preliminary “perception-reasoning-action” loop in complex urban environments.  

The Aerospace Embodied Task Planning task is positioned as the advanced planning stage in the curriculum. This task requires the drone to generate detailed path plans (e.g., “\textit{First go straight to the red building, then turn right and bypass the tall building}”) based on the starting and ending points. This stage is an extension of the navigation task but places greater emphasis on global planning capabilities.

\subsection{SkyAgent \& SkyAgentX}
As illustrated in Figure \ref{FIG:model_framework}, based on a pre-trained general vision-language model Internvl-2.5-8B\cite{chen2024internvl}, we introduce the aerospace embodied chain-of-thought mechanism and fine-tune it using specialized instruction datasets (i.e., SkyAgent-Scene3k, SkyAgent-Reason3k, SkyAgent-Nav3k, and SkyAgent-Plan3k) for four downstream tasks, to develop the SkyAgent model. On this foundation, we further incorporate the aerospace embodied multitask curriculum learning strategy, progressively training the model through multi-task joint training in the sequential order of scene perception, spatial reasoning, navigation exploration, and task planning. This process ultimately led to the construction of the UAV-agent embodied large model, SkyAgentX, which integrates “perception-reasoning-navigating-planning” into an unified framework.

\section{Results}
\subsection{Quantitative Analysis}

\begin{table*}[hbt]
\centering
\caption{Evaluation results on the shanghai city test dataset:SkyAgent-Scene3k, SCENE refers to LLM-JUDGE-SCENE.}
\label{task_1_result}
\begin{tabular*}{1\linewidth}{c|ccc|ccc|ccc|ccc}
\toprule
\textbf{City$\rightarrow$}&\multicolumn{3}{c|}{ShangHai} &\multicolumn{3}{c|}{ShenZhen} &\multicolumn{3}{c|}{Campus} &\multicolumn{3}{c}{Residence} \\
\midrule
\textbf{Models$\downarrow$} & BLEU & SPICE &  SCENE& BLEU & SPICE & SCENE & BLEU & SPICE & SCENE & BLEU & SPICE & SCENE \\
\midrule
3d-llm\cite{Hong20233DLLMIT}&0.0346&0.0162&0.1660 &0.0283&0.0117&0.1156 &0.0429&0.0197&0.1511 &0.0335&0.0105&0.1378 \\
GPT-4-vision-preview\cite{Achiam2023GPT4TR} & 0.1200 & 0.0884 & \textbf{0.6840}  & 0.1039 & 0.0924 & 0.6909  & 0.1035 & 0.0901 & 0.6511  & 0.1246 & 0.0999 & \textbf{0.7444}  \\
GPT-4o\cite{Achiam2023GPT4TR} & 0.1539 & 0.1114 & 0.6800  & 0.1532 & 0.1277 & \textbf{0.7178}  & 0.1237 & 0.1175 & \textbf{0.6977}  & 0.1606 & 0.1225 & 0.7089  \\
blip2-flan-t5-xxl\cite{Li2023BLIP2BL} & 0.1954 & 0.0860 & 0.4041  & 0.1932 & 0.0956 & 0.4333  & 0.2151 & 0.0906 & 0.4318  & 0.2593 & 0.1307 & 0.5089  \\
blip2-opt-6.7b\cite{Li2023BLIP2BL} & 0.1968 & 0.0814 & 0.4201  & 0.2279 & 0.0836 & 0.4289  & 0.2140 & 0.0906 & 0.4533  & 0.2558 & 0.1102 & 0.4778  \\
instructblip-flan-t5-xxl\cite{Dai2023InstructBLIPTG}& 0.2118 & 0.0808 & 0.4908  & 0.1972 & 0.0852 & 0.4689  & 0.2149 & 0.0969 & 0.5067  & 0.2536 & 0.1202 & 0.5400  \\
instructblip-vicuna-7b\cite{Dai2023InstructBLIPTG}& 0.2239 & 0.0787 & 0.4911  & 0.2102 & 0.0835 & 0.5022  & 0.2109 & 0.0867 & 0.4889  & 0.2729 & 0.1193 & 0.5511 \\
instructblip-vicuna-13b\cite{Dai2023InstructBLIPTG}& 0.2185 & 0.0810 & 0.4752  & 0.2176 & 0.0852 & 0.4533  & 0.2161 & 0.0832 & 0.4556  & 0.2715 & 0.1084 & 0.5644  \\
llama-adapter-v2-7B\cite{Gao2023LLaMAAdapterVP}& 0.0843 & 0.0512 & 0.5236  & 0.0741 & 0.0546 & 0.5067  & 0.0730 & 0.0584 & 0.5378  & 0.0981 & 0.0715 & 0.5778  \\
llava-v1.5-vicuna-7b\cite{Liu2023ImprovedBW}& 0.0746 & 0.0469 & 0.5000  & 0.0639 & 0.0515 & 0.5364  & 0.0590 & 0.0533 & 0.5133  & 0.0790 & 0.0645 & 0.5933  \\
llava-v1.5-vicuna-13b\cite{Liu2023ImprovedBW}& 0.0731 & 0.0468 & 0.5314  & 0.0643 & 0.0545 & 0.5727  & 0.0604 & 0.0505 & 0.5511  & 0.0754 & 0.0673 & 0.6067  \\
llava-v1.6-vicuna-7b\cite{Liu2023ImprovedBW}&0.0483 & 0.0019 & 0.4823  &0.0423 & 0.0025 & 0.5289  &0.0387 & 0.0036 & 0.5178  &0.0545 & 0.0148 & 0.5778  \\
llava-v1.6-vicuna-13b\cite{Liu2023ImprovedBW}&0.0484 & 0.0039 & 0.5010  &0.0417 & 0.0024 & 0.5364  &0.0395 & 0.0047 & 0.4738  &0.0525 & 0.0057 & 0.5489  \\
miniGPT4\cite{Zhu2023MiniGPT4EV}&0.0969 & 0.0613 & 0.5592  &0.0824 & 0.0584 & 0.5467  &0.0787 & 0.0605 & 0.4607  &0.0801 & 0.0624 & 0.4489  \\
mplug\cite{Ye2023mPLUGOwlME}\cite{Ye2023mPLUGOwlME}&0.0605 & 0.0450 & 0.5626  &0.0520 & 0.0490 & 0.5533  &0.0522 & 0.0489 & 0.5585  &0.0680 & 0.0582 & 0.5400  \\
mplug2\cite{Ye2023mPLUGOwl2RM}&0.0928 & 0.0502 & 0.5276  &0.0825 & 0.0590 & 0.5796  &0.0675 & 0.0447 & 0.5705  &0.1020 & 0.0714 & 0.5614  \\
qwen-lv-7b\cite{Bai2023QwenVLAV}&0.2305 & 0.0946 & 0.4780  &0.2417 & 0.1136 & 0.5133  &0.2206 & 0.0946 & 0.4467  &0.2682 & 0.1057 & 0.4733 \\
\midrule
\textbf{SkyAgent (ours)} & \textbf{0.4302}&	\textbf{0.3083} &	0.5166&	\textbf{0.5349}&	\textbf{0.4068}&	0.4831 &	\textbf{0.5085}&	\textbf{0.3349}&	0.4732&	\textbf{0.4278}&	\textbf{0.2557}&	0.5295 \\

\bottomrule
\end{tabular*}
\end{table*}
%task_2 shanghai
\begin{table*}[hbt]
\centering
\caption{Evaluation results on the shanghai city test dataset:SkyAgent-Reason3k, Rea refers to LLM-JUDGE-Reason.}
\label{task_2_result}
\begin{tabular*}{1\linewidth}{c|ccc|ccc|ccc|ccc}
\toprule
\textbf{City$\rightarrow$}&\multicolumn{3}{c|}{ShangHai} &\multicolumn{3}{c|}{ShenZhen} &\multicolumn{3}{c|}{Campus} &\multicolumn{3}{c}{Residence} \\
\midrule
\textbf{Models$\downarrow$} & BLEU & SPICE &  REA& BLEU & SPICE &REA & BLEU & SPICE &REA & BLEU & SPICE &REA \\
\midrule
3d-llm\cite{Hong20233DLLMIT}&0.1310&0.1008&0.3180 &0.1839&0.1305&0.3133 &0.0532&0.0373& 0.1778 &0.0792&0.009&0.2889 \\
GPT-4-vision-preview\cite{Achiam2023GPT4TR} & 0.0696 & 0.0701 & 0.3680  & 0.0830 & 0.1233 & 0.4578  & 0.0261 & 0.0154 & 0.3600  & 0.0917 & 0.1064 & 0.2822  \\
GPT-4o\cite{Achiam2023GPT4TR} & 0.1498 & 0.1710 & \textbf{0.493}  & 0.1809 & 0.2034 & \textbf{0.4733}  & 0.0558 & 0.0608 & 0.4467  & 0.3213 & 0.3750 & 0.4844  \\
blip2-flan-t5-xxl\cite{Li2023BLIP2BL}& 0.0661 & 0.0863 & 0.3387  & 0.0867 & 0.1252 & 0.2756  & 0.0174 & 0.0089 & 0.1978  & 0.0868 & 0.0677 & 0.3844  \\
blip2-opt-6.7b\cite{Li2023BLIP2BL}& 0.0508 & 0.0685 & 0.2023  & 0.0452 & 0.0804 & 0.1444  & 0.0533 & 0.0405 & 0.2089  & 0.0548 & 0.1619 & 0.2156  \\
instructblip-flan-t5-xxl\cite{Dai2023InstructBLIPTG}& 0.0966 & 0.1207 & 0.3590  & 0.1351 & 0.1725 & 0.2556  & 0.0354 & 0.0296 & 0.2133  & 0.1261 & 0.0857 & 0.3800  \\
instructblip-vicuna-7b\cite{Dai2023InstructBLIPTG}& 0.0254 & 0.0393 & 0.2630  & 0.0207 & 0.0493 & 0.2244  & 0.0480 & 0.0649 & 0.2978  & 0.0865&0.1088 & 0.2667  \\
instructblip-vicuna-13b\cite{Dai2023InstructBLIPTG}& 0.0158 & 0.0116 & 0.3620  & 0.0260 & 0.0278 & 0.2867  & 0.0041 & 0 & 0.1978  & 0.0002&0 & 0.2800  \\
llama-adapter-v2-7B\cite{Gao2023LLaMAAdapterVP}& 0.1582 & 0.1792 & 0.3430  & 0.1720 & 0.2164 & 0.3822  & 0.0721 & 0.0702 & 0.2422  & 0.3137&0.4432 & \textbf{0.5068}  \\
llava-v1.5-vicuna-7b\cite{Liu2023ImprovedBW}& 0.1054 & 0.1269 & 0.3380  & 0.1190 & 0.1587 & 0.3046  & 0.0422 & 0.0412 & 0.4000  & 0.2339&0.3033 & 0.3667  \\
llava-v1.5-vicuna-13b\cite{Liu2023ImprovedBW}& 0.1235 & 0.1386 & 0.3760  & 0.1205 & 0.1837 & 0.3489  & 0.0509 & 0.0473 & 0.3911  & 0.2159&0.2779 & 0.3600  \\
llava-v1.6-vicuna-7b\cite{Liu2023ImprovedBW}&0.0653 & 0.0887 & 0.3020  &0.1016 & 0.1517 & 0.3444  &0.0214 & 0.0196 & 0.3046  &0.1123&0.1417 & 0.2733  \\
llava-v1.6-vicuna-13b\cite{Liu2023ImprovedBW}&0.0680 & 0.0969 & 0.3490  &0.0731 & 0.1176 & 0.3178  &0.0250 & 0.0275 & 03378  &0.106&0.164 & 0.3556  \\
miniGPT4\cite{Zhu2023MiniGPT4EV}&0.1307 & 0.1714 & 0.3930  &0.1211 & 0.1895 & 0.260  &0.0288 & 0.0306 & 0.3022  &0.1784&0.2266 & 0.3556  \\
mplug\cite{Ye2023mPLUGOwlME}&0.1277 & 0.1436 & 0.313  &0.1551 & 0.1932 & 0.3133  &0.0659 & 0.0770 & 0.3711  &0.2356&0.3148 & 0.3156  \\
mplug2\cite{Ye2023mPLUGOwl2RM}&0.1375 & 0.1303 & 0.373  &0.1468 & 0.1444 & 0.3288  &0.0649 & 0.0520 & 0.3800  &0.2668&0.238 & 0.4222  \\
qwen-lv-7b\cite{Bai2023QwenVLAV}&0.1310 & 0.1590 & 0.305  &0.1475 & 0.1878 & 0.2932  &0.0873 & 0.0719 & \textbf{0.6244}  &0.2432&0.3324 & 0.3796 \\
\midrule
\textbf{SkyAgent (ours)} & 0.4598&	0.3846 &	0.4310&	0.3552&	\textbf{0.3255}&	0.2729 &	0.3323&	0.1648&	0.5792&	0.3570&	0.3739&	0.3189 \\

\textbf{SkyAgentX (ours)} &\textbf{0.4660}&	\textbf{0.4185} &	\textbf{0.5309}&	\textbf{0.3601}&	0.3195&	0.3295 &	\textbf{0.3469}&	\textbf{0.1777}&	0.6043&	\textbf{0.4399}&	\textbf{0.4439}&	0.4197 \\
\bottomrule
\end{tabular*}
\end{table*}

%task_3 shanghai
\begin{table*}[hbt]
\centering
\caption{Evaluation results on the shanghai city test dataset:SkyAgent-Nav3k, Nav refers to LLM-JUDGE-NAV.}
\label{task_3_result}
\begin{tabular*}{1\linewidth}{c|ccc|ccc|ccc|ccc}
\toprule
\textbf{City$\rightarrow$}&\multicolumn{3}{c|}{ShangHai} &\multicolumn{3}{c|}{ShenZhen} &\multicolumn{3}{c|}{Campus} &\multicolumn{3}{c}{Residence} \\
\midrule
\textbf{Models$\downarrow$} & BLEU & SPICE &  NAV& BLEU & SPICE & NAV& BLEU & SPICE & NAV& BLEU & SPICE &NAV \\
\midrule
3d-llm\cite{Hong20233DLLMIT}&0.108&0.0312&0.2808 &0.0851&0.0081&0.2171 &0.1263&0.0&0.292 &0.0609&0.0425&0.2135 \\
GPT-4-vision-preview\cite{Achiam2023GPT4TR} & 0.1277&0.1871 & 0.3263  & 0.1193&0.1514 & 0.3643  & 0.049&0.0529 & 0.3320  & 0.0718&0.0836 & 0.3392  \\
GPT-4o\cite{Achiam2023GPT4TR} & 0.2343&0.2861 & 0.4741  & 0.2137&0.2004 & 0.3714  & 0.1349&0.1289 & \textbf{0.5960}  & 0.1473&0.1684& \textbf{0.4519}  \\
blip2-flan-t5-xxl\cite{Li2023BLIP2BL}& 0.080&0.0611 & 0.383  & 0.0686&0.0671 & 0.3357  & 0.0788&0.064 & 0.4100  & 0.1024&0.0433& 0.2789  \\
blip2-opt-6.7b\cite{Li2023BLIP2BL}& 0.0245&0.0228 & 0.1400  & 0.0092&0.004 & 0.1333  & 0.0221&0.018 & 0.1939  & 0.0358&0.0335 & 0.1789  \\
instructblip-flan-t5-xxl\cite{Dai2023InstructBLIPTG}& 0.0345&0.0248 & 0.3430  & 0.0324&0.0427 & 0.2845  & 0.0207&0.0367 & 0.4720  & 0.0637&0.0213& 0.3462  \\
instructblip-vicuna-7b\cite{Dai2023InstructBLIPTG}& 0.1844&0.2168 & 0.2350  & 0.1812&0.1576 & 0.2231  & 0.1187&0.09 & 0.1674  & 0.1364&0.155 & 0.2000  \\
instructblip-vicuna-13b\cite{Dai2023InstructBLIPTG}& 0.1917&0.2146 & 0.2855  & 0.1952&0.1874 & 0.3175  & 0.1245&0.1028 & 0.2800  & 0.1508&0.1963& 0.2765  \\
llama-adapter-v2-7B\cite{Gao2023LLaMAAdapterVP}& 0.2324&0.2692 & 0.3829  & 0.2067&0.1976 & 0.3000  & 0.1366&0.1191 & 0.4348  & 0.1789&0.194& 0.3217  \\
llava-v1.5-vicuna-7b\cite{Liu2023ImprovedBW}& 0.1868&0.229 & 0.3688  & 0.1461&0.1575 & 0.2786  & 0.0727&0.0762 & 0.4660  & 0.1297&0.1274 & 0.3500  \\
llava-v1.5-vicuna-13b\cite{Liu2023ImprovedBW}& 0.1809&0.218 & 0.3309  & 0.1701&0.1821 & 0.3500  & 0.0971&0.099 & 0.4490  & 0.142&0.1532 & 0.3173  \\
llava-v1.6-vicuna-7b\cite{Liu2023ImprovedBW}&0.1372&0.1977 & 0.3242  &0.1171&0.1397 & 0.3571  &0.0786&0.0936 & 0.3939  &0.075&0.0977 & 0.2962  \\
llava-v1.6-vicuna-13b\cite{Liu2023ImprovedBW}&0.1251&0.1743 & 0.3503  &0.1099&0.1439 & 0.2952  &0.0607&0.060 & 0.4020  &0.0761&0.1002 & 0.2865  \\
miniGPT4\cite{Zhu2023MiniGPT4EV}&0.2199&0.2969 & 0.4080  &0.1896&0.1978 & 0.3310  &0.1188&0.120 & 0.5660  &0.1374&0.1805 & 0.3750  \\
mplug\cite{Ye2023mPLUGOwlME}&0.1971&0.2485 & 0.3882  &0.1712&0.1729 & 0.2833  &0.0773&0.0842 & 0.3960  &0.0963&0.1248 & 0.3000  \\
mplug2\cite{Ye2023mPLUGOwl2RM}&0.2127&0.2081 & 0.4121  &0.1881&0.1411 & 0.3643  &0.1125&0.0935 & 0.4900  &0.1474&0.1349& 0.3404  \\
qwen-lv-7b\cite{Bai2023QwenVLAV}&0.1772&0.2152 & 0.3505  &0.1941&0.1837 & 0.3857  &0.1267&0.1278 & 0.5225  &0.1496&0.1128& 0.3827 \\
\midrule
\textbf{SkyAgent (ours)}&0.3660&	0.3535&	\textbf{0.4903}&	\textbf{0.3519}&	\textbf{0.2947}&	0.4387&	0.2962&	0.2813	&0.5805	&0.2961	&\textbf{0.2329}&	0.3189 \\

\textbf{SkyAgentX (ours)}&\textbf{0.3669}&	\textbf{0.3590}&	0.4741&	0.2976&	0.2912&	\textbf{0.4531}&	\textbf{0.3466}&	\textbf{0.3517}	&0.5842	&\textbf{0.2253}	&0.1763&	0.3138 \\
\bottomrule
\end{tabular*}
\end{table*}

%task_4 shanghai
\begin{table*}[hbt]
\centering
\caption{Evaluation results on the shanghai city test dataset:SkyAgent-Plan3k, PLAN refers to LLM-JUDGE-PLAN.}
\label{task_4_result}
\begin{tabular*}{1\linewidth}{c|ccc|ccc|ccc|ccc}
\toprule
\textbf{City$\rightarrow$}&\multicolumn{3}{c|}{ShangHai} &\multicolumn{3}{c|}{ShenZhen} &\multicolumn{3}{c|}{Campus} &\multicolumn{3}{c}{Residence} \\
\midrule
\textbf{Models$\downarrow$} & BLEU & SPICE &  PLAN& BLEU & SPICE & PLAN & BLEU & SPICE & PLAN & BLEU & SPICE &PLAN \\
\midrule
3d-llm\cite{Hong20233DLLMIT}&0.0002&0.0019&0.1440 &0.0014&0.0027&0.1000 &0.0023&0.0021&0.1400 &0.0&0.0&0.1044 \\
GPT-4-vision-preview\cite{Achiam2023GPT4TR} &0.1073&0.0457& 0.4520  &0.1269&0.0497& 0.4067  &0.1231&0.0426& 0.3711  &0.1168&0.0445& 0.3867  \\
GPT-4o\cite{Achiam2023GPT4TR} &0.1064&0.0512& 0.5290  &0.1152&0.0520& 0.5077  &0.1167&0.0503& 0.4750  &0.1183&0.0483& 0.5178  \\
blip2-flan-t5-xxl\cite{Li2023BLIP2BL}& 0.0001&0.0566& 0.2000  & 0.0001&0.0408& 0.1844  & 0.0016&0.0537& 0.2200  & 0.0024&0.0482& 0.1956  \\
blip2-opt-6.7b\cite{Li2023BLIP2BL}& 0.0&0.0002& 0.1000  & 0.0&0.0 & 0.1000  & 0.0&0.0 & 0.1000  & 0.0&0.0026& 0.1044  \\
instructblip-flan-t5-xxl\cite{Dai2023InstructBLIPTG}& 0.0004&0.0579& 0.2356  & 0.0003&0.0423& 0.1875  & 0.0095&0.0458& 0.2733  & 0.0014&0.072& 0.2667  \\
instructblip-vicuna-7b\cite{Dai2023InstructBLIPTG}& 0.0015&0.0068& 0.1386  & 0.0004&0.0323& 0.1546  & 0.0074&0.0121& 0.1667  & 0.0018&0.0186& 0.1625  \\
instructblip-vicuna-13b\cite{Dai2023InstructBLIPTG}& 0.0012&0.0114& 0.1610  & 0.0034&0.0218& 0.1579  & 0.0063&0.0046& 0.1452  & 0.0007&0.0151& 0.1667  \\
llama-adapter-v2-7B\cite{Gao2023LLaMAAdapterVP}& 0.0349&0.0216& 0.1905  & 0.0003&0.0& 0.1500  & 0.1138&0.0381& 0.2444  & 0.0838&0.0195& 0.1800  \\
llava-v1.5-vicuna-7b\cite{Liu2023ImprovedBW}& 0.1324&0.0339& 0.3212  & 0.1654&0.0357& 0.3289  & 0.1662&0.0376& 0.3267  & 0.1603&0.0414& 0.3111  \\
llava-v1.5-vicuna-13b\cite{Liu2023ImprovedBW}& 0.1375&0.0355& 0.3640  & 0.1612&0.038& 0.3200  & 0.1622&0.0448& 0.3600  & 0.1489&0.0497& 0.3022  \\
llava-v1.6-vicuna-7b\cite{Liu2023ImprovedBW}&0.1420&0.0406& 03590  &0.1697&0.044&0.3756&0.1442&0.0382& 0.3727  &0.1618&0.042& 0.3533  \\
llava-v1.6-vicuna-13b\cite{Liu2023ImprovedBW}&0.1246&0.0399& 0.3250  &0.1388&0.0391& 0.3089  &0.1378&0.034& 0.3067  &0.1535&0.0452& 0.3400  \\
miniGPT4\cite{Zhu2023MiniGPT4EV}&0.1288&0.0403& 0.2770  &0.1396&0.0451& 0.2511  &0.1689&0.0425& 0.2511  &0.1585&0.0451& 0.2622  \\
mplug\cite{Ye2023mPLUGOwlME}&0.1368&0.0368& 0.3020  &0.1661&0.0397& 0.3136  &0.1607&0.0393& 0.3400  &0.1786&0.0409& 0.3244  \\
mplug2\cite{Ye2023mPLUGOwl2RM}&0.1426&0.0343& 0.3230  &0.1687&0.0392& 0.3378  &0.1674&0.0379& 03114  &0.1673&0.0442& 0.2978  \\
qwen-lv-7b\cite{Bai2023QwenVLAV}&0.1574&0.0541& 0.2850  &0.1536&0.0506& 0.2956  &0.1218&0.0310& 0.2800  &0.1001&0.0375& 0.2178  \\
\midrule
\textbf{SkyAgent (ours)}&0.4087&	\textbf{0.2653}&	0.6175&	\textbf{0.3605}&	0.2404&	0.6016	&\textbf{0.3403}&	0.2257&	0.6441&	\textbf{0.4256}&	0.2429&	0.5821 \\

\textbf{SkyAgentX (ours)}&\textbf{0.4230}&	0.2557&	\textbf{0.6217}&	0.3598&	\textbf{0.2704}&	\textbf{0.6083}	&0.3055&	\textbf{0.2327}&	\textbf{0.6457}&	0.3959&	\textbf{0.2753}&	\textbf{0.5885} \\
\bottomrule
\end{tabular*}
\end{table*}

\begin{table*}[hbt]
% \small
\footnotesize
\centering
\caption{Consolidated Human Evaluation scores across all models, tasks, and scenarios. Scores are on a scale of 0 to 1. Abbreviations for scenarios are as follows: SH (ShangHai), SZ (ShenZhen), CA (Campus), and RE (Residence).}
\label{human_eval_summary}
\begingroup
\setlength{\tabcolsep}{4pt}
\resizebox{\textwidth}{!}{
\begin{tabular}{l|cccc|cccc|cccc|cccc}
\toprule

\multicolumn{1}{c|}{} & \multicolumn{4}{c|}{SkyAgent-Scene3k} & \multicolumn{4}{c|}{SkyAgent-Reason3k} & \multicolumn{4}{c|}{SkyAgent-Nav3k} & \multicolumn{4}{c}{SkyAgent-Plan3k} \\
\cmidrule{2-17}

Models$\downarrow$ &
SH & SZ & CA & RE &
SH & SZ & CA & RE &
SH & SZ & CA & RE &
SH & SZ & CA & RE \\
\midrule

3d-llm\cite{Hong20233DLLMIT} & 0.12 & 0.11 & 0.14 & 0.13 & 0.37 & 0.60 & 0.23 & 0.25 & 0.34 & 0.29 & 0.40 & 0.25 & 0.10 & 0.11 & 0.12 & 0.10 \\
gpt-4-vision-preview\cite{Achiam2023GPT4TR} & 0.41 & 0.32 & 0.30 & 0.40 & 0.20 & 0.30 & 0.16 & 0.29 & 0.40 & 0.39 & 0.19 & 0.30 & 0.36 & 0.43 & 0.46 & 0.38 \\
gpt-4o\cite{Achiam2023GPT4TR} & 0.45 & 0.38 & 0.35 & 0.48 & 0.40 & 0.59 & 0.22 & 0.75 & 0.70 & 0.68 & 0.44 & 0.57 & 0.35 & 0.41 & 0.45 & 0.39 \\
blip2-flan-t5-xxl\cite{Li2023BLIP2BL} & 0.53 & 0.46 & 0.50 & 0.68 & 0.19 & 0.31 & 0.14 & 0.28 & 0.28 & 0.25 & 0.30 & 0.41 & 0.10 & 0.11 & 0.12 & 0.11 \\
blip2-opt-6.7b\cite{Li2023BLIP2BL} & 0.54 & 0.51 & 0.49 & 0.67 & 0.17 & 0.20 & 0.21 & 0.19 & 0.14 & 0.12 & 0.13 & 0.18 & 0.10 & 0.10 & 0.10 & 0.10 \\
instructblip-flan-t5-xxl\cite{Dai2023InstructBLIPTG} & 0.58 & 0.47 & 0.51 & 0.66 & 0.28 & 0.45 & 0.18 & 0.38 & 0.17 & 0.17 & 0.15 & 0.27 & 0.11 & 0.11 & 0.14 & 0.11 \\
instructblip-vicuna-7b\cite{Dai2023InstructBLIPTG} & 0.61 & 0.49 & 0.48 & 0.72 & 0.12 & 0.14 & 0.20 & 0.27 & 0.58 & 0.59 & 0.38 & 0.52 & 0.12 & 0.11 & 0.13 & 0.11 \\
instructblip-vicuna-13b\cite{Dai2023InstructBLIPTG} & 0.60 & 0.50 & 0.52 & 0.71 & 0.10 & 0.15 & 0.11 & 0.10 & 0.60 & 0.63 & 0.39 & 0.59 & 0.11 & 0.12 & 0.13 & 0.10 \\
llama-adapter-v2-7B\cite{Gao2023LLaMAAdapterVP} & 0.19 & 0.17 & 0.21 & 0.25 & 0.41 & 0.57 & 0.29 & 0.74 & 0.69 & 0.66 & 0.45 & 0.65 & 0.20 & 0.11 & 0.44 & 0.30 \\
llava-v1.5-vicuna-7b\cite{Liu2023ImprovedBW} & 0.18 & 0.16 & 0.17 & 0.22 & 0.30 & 0.40 & 0.19 & 0.59 & 0.59 & 0.49 & 0.28 & 0.50 & 0.40 & 0.53 & 0.52 & 0.47 \\
llava-v1.5-vicuna-13b\cite{Liu2023ImprovedBW} & 0.17 & 0.15 & 0.18 & 0.21 & 0.34 & 0.41 & 0.21 & 0.55 & 0.57 & 0.56 & 0.34 & 0.55 & 0.42 & 0.52 & 0.54 & 0.45 \\
llava-v1.6-vicuna-7b\cite{Liu2023ImprovedBW} & 0.14 & 0.13 & 0.13 & 0.17 & 0.18 & 0.35 & 0.15 & 0.34 & 0.43 & 0.38 & 0.29 & 0.32 & 0.42 & 0.57 & 0.49 & 0.48 \\
llava-v1.6-vicuna-13b\cite{Liu2023ImprovedBW} & 0.15 & 0.12 & 0.12 & 0.16 & 0.19 & 0.28 & 0.15 & 0.32 & 0.39 & 0.36 & 0.24 & 0.33 & 0.38 & 0.46 & 0.48 & 0.46 \\
minigpt4\cite{Zhu2023MiniGPT4EV} & 0.21 & 0.20 & 0.23 & 0.23 & 0.35 & 0.42 & 0.17 & 0.49 & 0.65 & 0.61 & 0.38 & 0.53 & 0.39 & 0.47 & 0.59 & 0.47 \\
mplug\cite{Ye2023mPLUGOwlME} & 0.16 & 0.14 & 0.19 & 0.19 & 0.35 & 0.52 & 0.27 & 0.60 & 0.62 & 0.57 & 0.29 & 0.39 & 0.41 & 0.54 & 0.53 & 0.53 \\
mplug2\cite{Ye2023mPLUGOwl2RM} & 0.20 & 0.19 & 0.20 & 0.28 & 0.38 & 0.48 & 0.26 & 0.68 & 0.64 & 0.60 & 0.37 & 0.57 & 0.43 & 0.56 & 0.58 & 0.50 \\
qwen-lv-7b\cite{Bai2023QwenVLAV} & 0.63 & 0.55 & 0.52 & 0.70 & 0.36 & 0.49 & 0.33 & 0.62 & 0.55 & 0.62 & 0.42 & 0.58 & 0.46 & 0.50 & 0.46 & 0.34 \\
\midrule
\textbf{SkyAgent (ours)}& \textbf{0.87} & \textbf{0.89} & \textbf{0.88} & \textbf{0.86} & 0.86 & 0.84 & 0.81 & 0.82 & 0.84 & \textbf{0.84} & 0.82 & \textbf{0.81} & 0.88 & \textbf{0.89} & \textbf{0.87} & \textbf{0.89} \\
\textbf{SkyAgentX (ours)}& 0.86 & 0.85 & 0.87 & 0.82 & \textbf{0.88} & \textbf{0.85} & \textbf{0.83} & \textbf{0.89} & \textbf{0.85} & 0.80 & \textbf{0.86} & 0.76 & \textbf{0.90} & 0.88 & 0.84 & 0.86 \\

\bottomrule
\end{tabular}
}
\endgroup
\end{table*}

\renewcommand{\dblfloatpagefraction}{0.9}
\begin{figure*}[t]
	\centering
		\includegraphics[scale=.365]{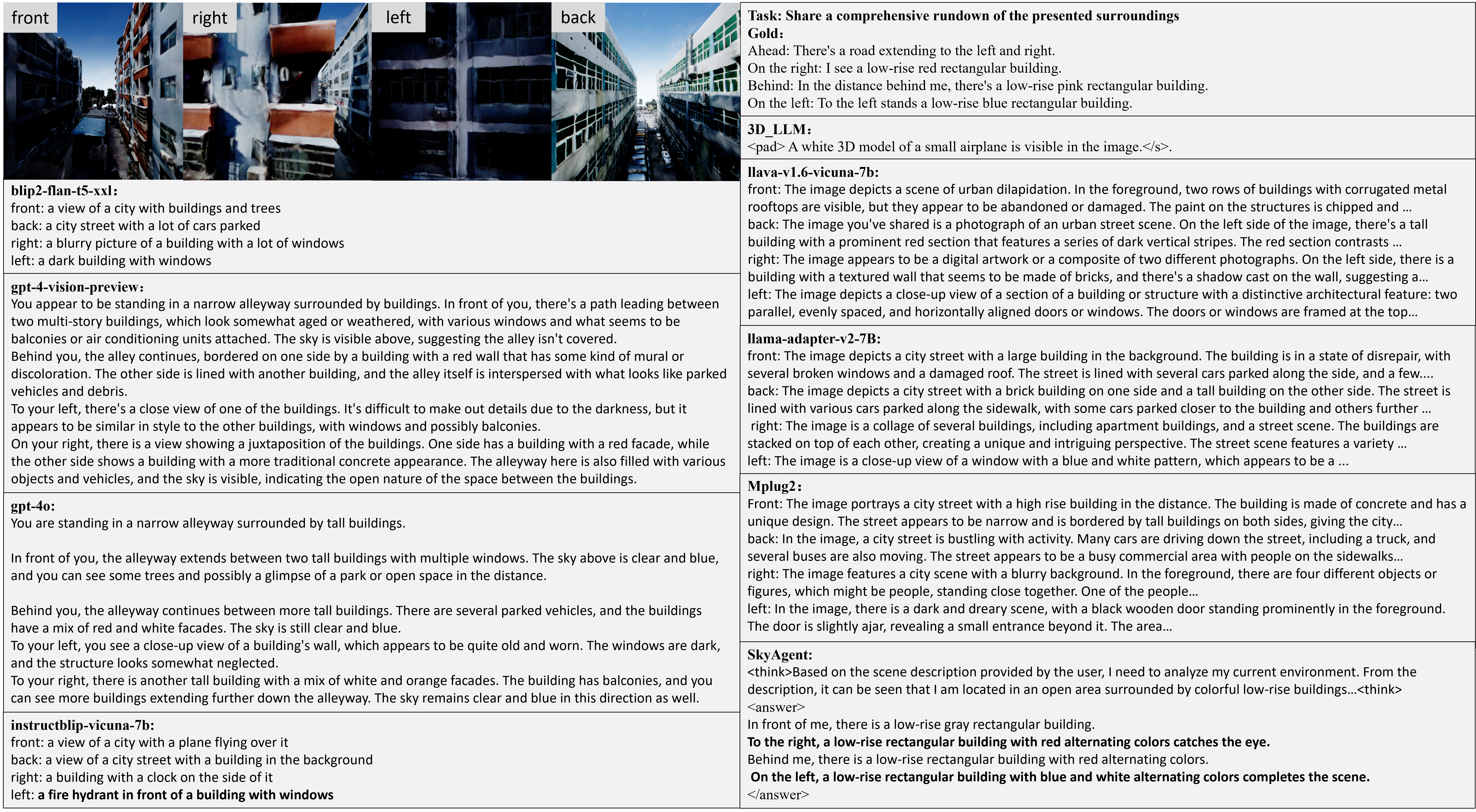}
	\caption{An example in the SkyAgent-Scene3k dataset.
 }
	\label{FIG:SpaceAgent_task_1_case}
\end{figure*}
\renewcommand{\dblfloatpagefraction}{0.9}
\begin{figure*}[t]
	\centering
		\includegraphics[scale=.4]{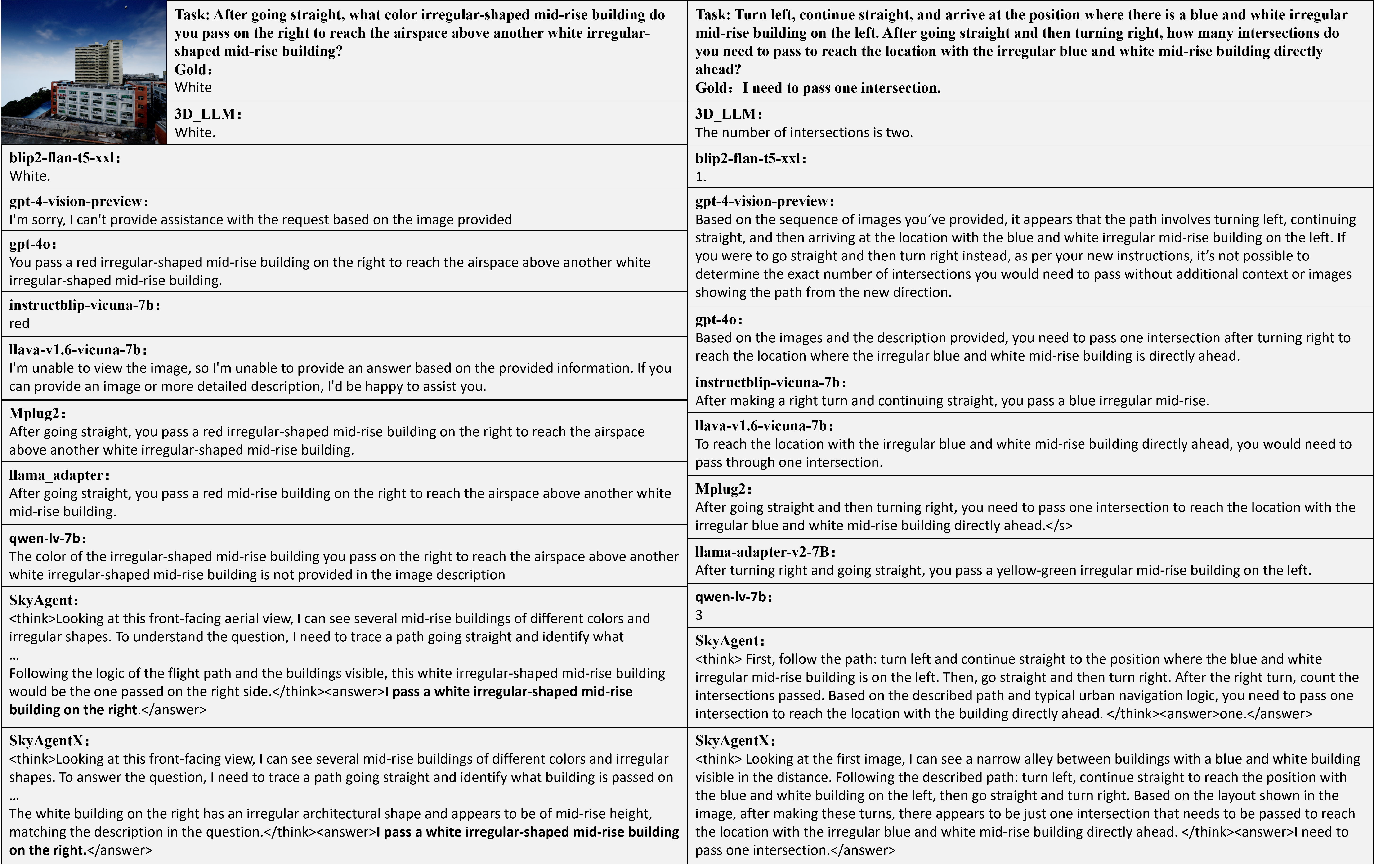}
	\caption{The examples in the SkyAgent-Reason3k dataset (left) and SkyAgent-Nav3k dataset (right).
 }
	\label{FIG:SpaceAgent_task_2_and_3}
\end{figure*}
As presented in Tables \ref{task_1_result}, \ref{task_2_result}, \ref{task_3_result}, and \ref{task_4_result}, we summarize the overall performance of visual-language models across four UAV downstream tasks within the AeroVerse benchmark. Despite significant advancements in both 2D and 3D visual-language models (VLMs) in recent years, these models continue to encounter challenges with UAV-agent embodied tasks, including the GPT-4 series. Among the four tasks, existing visual-language models achieve relatively high scores only on SkyAgent-Scene3k, while their performance on the other tasks declines markedly. Overall, GPT-4-vision-review and GPT-4o consistently outperform other models. However, our SkyAgentX trained by aerospace embodied chain-of-thought and multitask curriculum learning mechanisms has demonstrated breakthrough improvements across multiple metrics. To further validate these results, we also conduct a comprehensive human evaluation, detailed in Table \ref{human_eval_summary}, which confirms that outputs from our models are overwhelmingly preferred by human judges. We will subsequently provide a detailed analysis of the various embodied tasks. 

\textbf{Results on SkyAgent-Scene3k}. 
The evaluation of lexical richness, semantic accuracy, and human preference is conducted using BLEU, SPICE, and LLM-JUDGE-SCENE metrics. Qwen-lv-7b performes exceptionally well in BLEU, while GPT-4o led in SPICE, demonstrating its advantage in semantic matching. Notably, SkyAgent achieves a leapfrog improvement in both BLEU (Shanghai 0.4302 vs. Qwen 0.2305) and SPICE (0.3083 vs. GPT-4o 0.1114), indicating that the aerospace embodied chain-of-thought mechanism significantly enhances lexical and semantic expression capabilities. However, its LLM-JUDGE-SCENE score remains slightly lower than the GPT-4 series, suggesting that further optimization is needed for human preference alignment. Nevertheless, our direct human evaluation tells a different story. As shown in Table \ref{human_eval_summary}, SkyAgent achieves a human preference score of 0.87 in the Shanghai scenario, far surpassing GPT-4o (0.45). This indicates that human evaluators strongly prefer the detailed and accurate descriptions generated by our model, a nuance not fully captured by the automated LLM-JUDGE metric.

\textbf{Results on SkyAgent-Reason3k}. For evaluating, we utilize LLM-JUDGE-REASON to assess human preferences. Three models emerge as prominent in this context, i.e., two open-source models, llama-adapter-v2-7B \cite{Gao2023LLaMAAdapterVP} and qwen-lv-7b \cite{Bai2023QwenVLAV}, along with one closed-source model, GPT-4o. A horizontal comparison among the GPT-4 series reveals that GPT-4o demonstrates superior capabilities in first-person spatial reasoning and question-answering tasks. SkyAgent achieves BLEU/SPICE scores of 0.4598/0.3846 in the Shanghai scenario, significantly outperforming the baseline model. After introducing aerospace embodied multitask curriculum learning, the REA score improves further, demonstrating that progressive learning enhances the model's understanding of complex spatial relationships. In the residence scenario, the curriculum learning variant (i.e., SkyAgentX) achieves a BLEU score of 0.4399, a 23\% increase over the original version (i.e., SkyAgent), highlighting its adaptability to local environmental characteristics. This quantitative superiority is strongly corroborated by our human evaluation (Table \ref{human_eval_summary}), where SkyAgentX achieves a score of 0.88 in Shanghai, decisively outperforming all baseline models, including GPT-4o (0.40), proving a significant human preference for our model's reasoning abilities.

\textbf{Results on SkyAgent-Nav3k}. Similarly to the evaluation metric with SkyAgent-Reason3k, we employ LLM-JUDGE-NAV to assess human preferences. In this task, GPT-4o demonstrates particularly strong performance, ranking at the top across most urban scenarios and evaluation metrics, including vocabulary level, semantic level, and human preference. In the residential area scenario, the response from llama-adapter-v2-7B \cite{Gao2023LLaMAAdapterVP} exhibit greater alignment with the correct answer at the lexical level compared to other models. SkyAgent achieves an NAV score of 0.4903 in the Shanghai navigation task, with its explicit reasoning capability being particularly suitable for problems involving spatial reasoning. Crucially, the human evaluation results in Table \ref{human_eval_summary} underscore our model's practical advantage. SkyAgentX scores 0.85 in Shanghai, significantly higher than GPT-4o’s 0.70. This suggests that the navigational instructions generated by our model are clearer, more accurate, and more useful to a human user, even when automated metrics show closer competition.

\textbf{Results on SkyAgent-Plan3k}. In evaluating this task, we utilize LLM-JUDGE-PLAN to assess human preferences. The performance of many models in this task is notably poor, with several indicators yielding a score of $0$. This deficiency arises from the necessity of acquiring a comprehensive range of environmental characteristics for planning long-distance paths. The 3D-LLM struggled to address our inquiries due to its limited generalization capabilities. Although we provide initial-view maps of the true path to the 2D-LLM as environmental information, either through image captions or multi-image input, this information remains incomplete. Through the aerospace embodied chain-of-thought mechanism of drone agent, SkyAgent achieves a PLAN score of 0.6175 in Shanghai tests, significantly surpassing GPT-4o. When combined with aerospace embodied multitask curriculum learning strategy, the planning capability of SkyAgentX improves further and delivers the best performance in the tests in four cities. This remarkable performance is overwhelmingly supported by human evaluations (Table \ref{human_eval_summary}). In the Shanghai scenario, SkyAgentX receives a score of 0.90 from human judges, highlighting its exceptional ability to generate coherent and viable long-range plans that are vastly superior to those from any other model, including GPT-4o (0.35). Thanks to the phased environmental complexity learning mechanism, the model demonstrates significantly enhanced stability in long-path planning for complex scenarios, providing key technical support for autonomous navigation exploration in real-world settings.

\subsection{Qualitative Analysis}

% \renewcommand{\dblfloatpagefraction}{0.9}
% \begin{figure}[t]
% 	\centering
% 		\includegraphics[scale=.4]{figures/SpaceAgent_task_3.png}
% 	\caption{An example in the SkyAgent-Nav3k dataset.
%  }
% 	\label{FIG:SpaceAgent_task_3}
% \end{figure}

\renewcommand{\dblfloatpagefraction}{0.9}
\begin{figure*}[t]
	\centering
		\includegraphics[scale=.4]{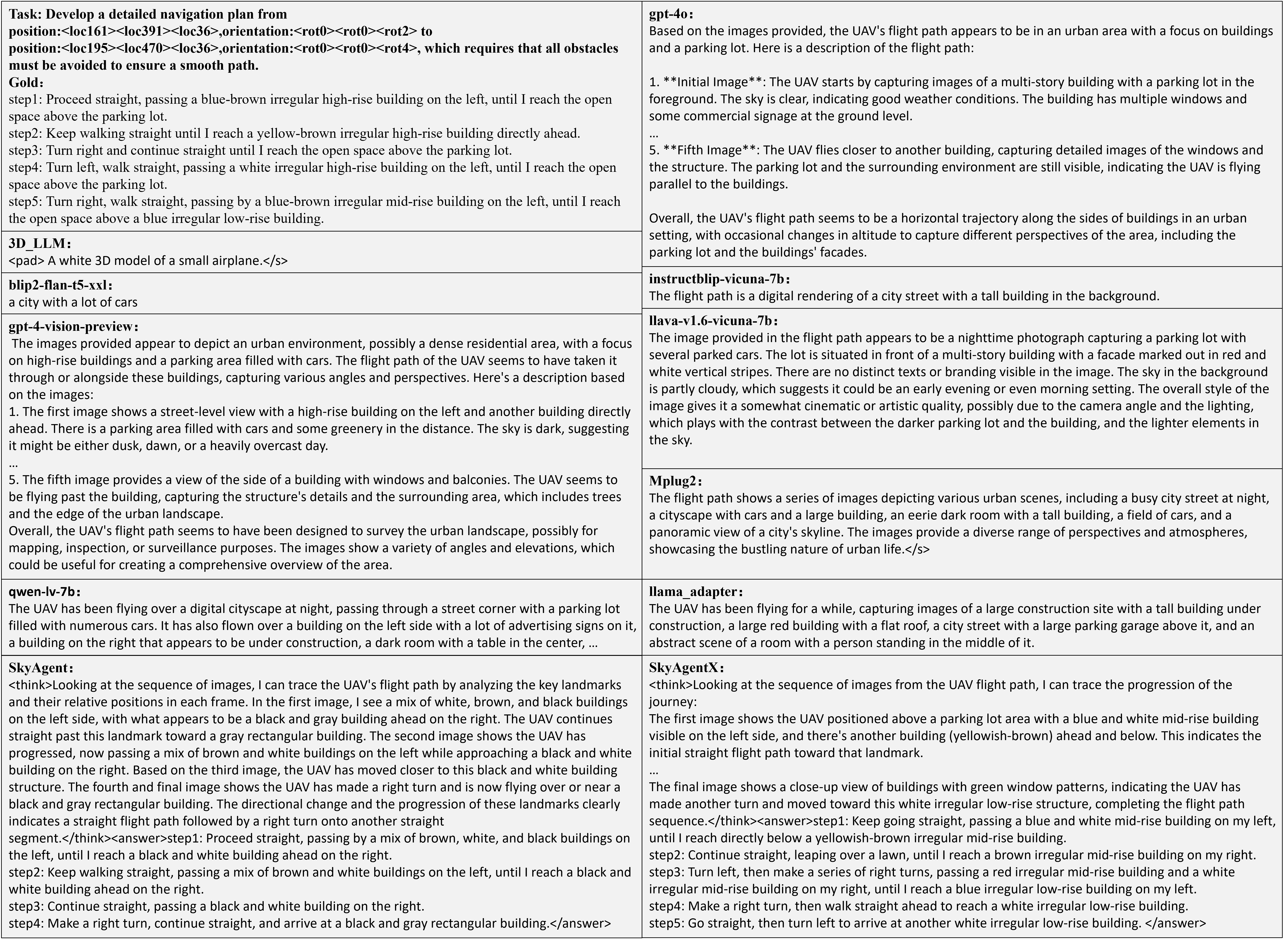}
	\caption{An example in the SkyAgent-Plan3k dataset.
 }
	\label{FIG:SpaceAgent_task_4}
\end{figure*}
\renewcommand{\dblfloatpagefraction}{0.9}
\begin{figure*}[t]
	\centering
		\includegraphics[width=\textwidth]{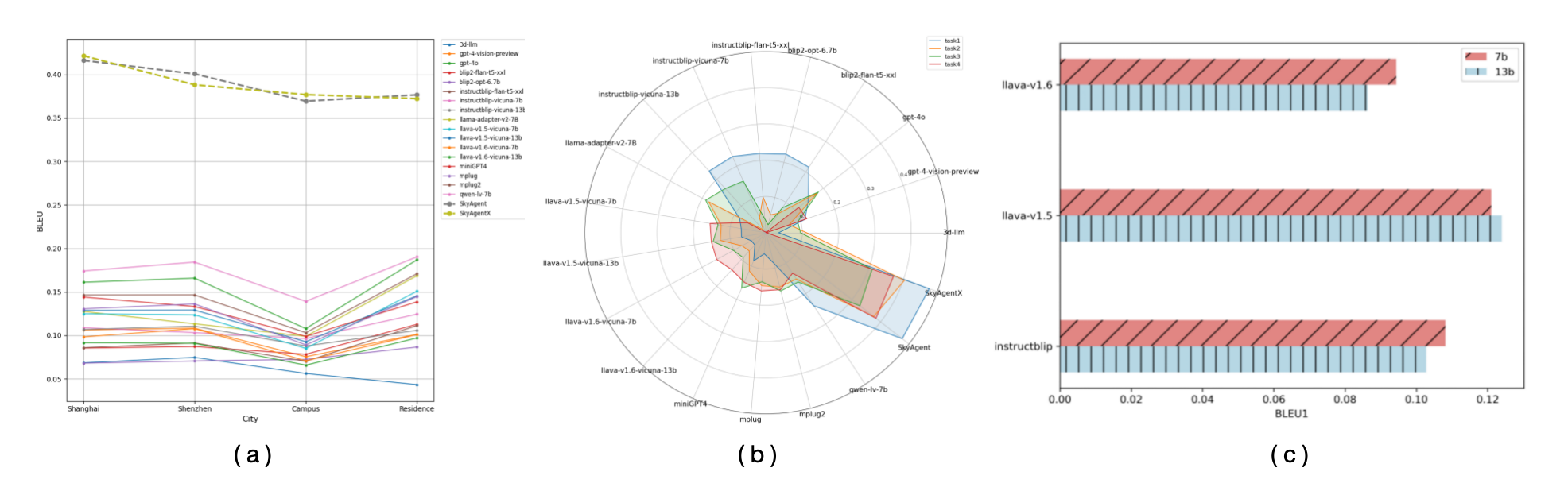}
	\caption{Model capability inquiry analysis diagram, where (a) is the scenario generalization capability diagram of the models, (b) is the task generalization capability diagram of the models, and (c) is the impact diagram of scaling law.
 }
	\label{FIG:Analyze}
\end{figure*}
%task_1
From Figure \ref{FIG:SpaceAgent_task_1_case}, although the 3D-LLM \cite{Hong20233DLLMIT} encodes the 3D environment and perceives its surroundings, it demonstrates limited generalization due to insufficient training on outdoor 3D urban data. When confronted with a 3D urban scene, the output of 3D-LLM\cite{Hong20233DLLMIT} resembles a description of an indoor environment, leading to significant hallucinations. The findings indicate that the performance of these 2D visual-language models surpasses that of 3D-LLM \cite{Hong20233DLLMIT}. This superiority can be attributed to the greater number of training image-text pairs available for the 2D visual-language models, which enhances their generalization capabilities. Furthermore, they deliver more accurate descriptions based on egocentric view images of urban settings. However, instances of hallucinations persist, such as the erroneous description of a fire hydrant in front of a building with windows in instructblip-vicuna-7b \cite{Dai2023InstructBLIPTG}, despite the absence of a fire hydrant in the image. Similarly, llama-adapter-v2-7B \cite{Gao2023LLaMAAdapterVP} inaccurately describes an individual walking in the distance, even though no person is present in the image. By contrast, SkyAgent’s results align best with the ground truth: the road ahead branches to the left and right; on the right is a low-rise red rectangular building, behind in the distance is a low-rise pink rectangular building, and on the left is a low-rise blue rectangular building. At the same time, it does not mischaracterize the outdoor scene as an indoor one like 3D‑LLM does.

%task_2
In the example illustrated in the left of Figure \ref{FIG:SpaceAgent_task_2_and_3}, the 3D visual-language model demonstrates the capability to perform short-term spatial reasoning based on its current posture and the 3D characteristics of the city to address questions. In contrast, the 2D visual-language model can derive answers solely from a single view image, which provides limited information. The results from this example indicate that the answer to the question is not present in the image. The building visible on the right reveals only a corner of the red roof in the bottom right of the image. This situation elucidates why 2D visual-language models, such as GPT-4o \cite{Achiam2023GPT4TR}, instructblip-vicuna-7b \cite{Dai2023InstructBLIPTG}, and qwen-lv-7b \cite{Bai2023QwenVLAV}, frequently reference red buildings or state their inability to provide a definitive answer. Conversely, the input of the 3D-LLM \cite{Hong20233DLLMIT} encompasses a more comprehensive and complex representation of urban features, leading to a correct and logical conclusion. In this case, both SkyAgent and SkyAgentX produce the correct answer, “white.” SkyAgent uses an embodied chain-of-thought to parse the constraints—“go straight, pass on the right, reach the airspace above the white building”—suppressing the distraction from the red roof and determining from relative orientation and path relations that the passed building is white. Building on this, SkyAgentX introduces multitask curriculum learning to align perception, reasoning, navigation, and planning, achieving more robust constraint propagation and a more concise explanation, clearly outperforming 2D models.

%task_3
In the example illustrated in the right of Figure \ref{FIG:SpaceAgent_task_2_and_3}, the model is tasked with following specific instructions to explore a distance before addressing subsequent questions. Due to the limitations of the 2D LLM, we derive its response to this question based on the parameters outlined in Section \ref{Baselines_Modification}, effectively simplifying the complexity of the question. The answer provided by the 3D-LLM \cite{Hong20233DLLMIT} is incorrect, as its input incorporates more complex 3D features. Currently, GPT-4-vision-preview and GPT-4o rank among the most advanced visual-language models, with GPT-4o demonstrating a slight advantage in addressing questions related to first-person view images. Instructblip-vicuna-7b \cite{Dai2023InstructBLIPTG} and llama-adapter-v2-7B \cite{Gao2023LLaMAAdapterVP} exhibit relatively weaker instruction-following capabilities, resulting in their inadequate responses to our inquiries. The responses from open-source 2D visual-language models, including blip2-flan-t5-xxl \cite{Li2023BLIP2BL}, llava-v1.6-vicuna-7b \cite{Liu2023ImprovedBW}, and Mplug2 \cite{Ye2023mPLUGOwl2RM}, are generally consistent with the gold standard, indicating their strong spatial reasoning and instruction-following abilities. In this case, SkyAgent outputs “One” which matches the ground truth. Its reasoning first anchors the blue-and-white irregular mid-rise building on the left, then infers via “turn left—go straight—turn right—count.” Benefiting from the aerospace embodied chain-of-thought and instruction fine-tuning, SkyAgent shows strong instruction-following ability, but references geometric details in first-person images insufficiently, which may cause ambiguity in complex alley scenes. With the gains from multitask curriculum learning, SkyAgentX is more robust under viewpoint changes and variations in target visibility, reducing hallucinations in intersection counting and spatial orientation, and overall performs better than mainstream 2D models.

%task_4
As illustrated in Figure \ref{FIG:SpaceAgent_task_4}, each model exhibits distinct responses to this task. The 3D-LLM’s capability for indoor task planning does not generalize effectively to urban environments, resulting in answers that are inconsistent with our queries and resembling 3D scene captions instead. To address the limitations of the 2D visual-language model, we derive its response to this question based on the parameters outlined in Section \ref{Baselines_Modification}. The Blip2-flan-t5-xxl model \cite{Li2023BLIP2BL} fails to accurately represent the flight path as per our instructions, instead providing an interpretation similar to image captioning, which indicates a relatively poor ability to adhere to directives. In contrast, both GPT-4o and GPT-4-vision-review \cite{Achiam2023GPT4TR} deliver the most detailed and comprehensive analyses of the initial views along the trajectory. The Instructblip-vicuna-7b \cite{Dai2023InstructBLIPTG}, llava-v1.6-vicuna-7b \cite{Liu2023ImprovedBW}, Mplug2 \cite{Ye2023mPLUGOwl2RM}, and llama-adapter-v2-7B \cite{Gao2023LLaMAAdapterVP} models do not describe the flight path in accordance with the timeline; instead, they provide a summary of the trajectory. SkyAgent can produce a step-wise flight path aligned with the timeline, anchoring the route using landmarks such as parking lot boundaries and black–white/blue–brown buildings, with markedly better instruction adherence than 2D/3D baselines. However, its semantic granularity is coarse, occasional left–right orientation drift occurs, and it does not explicitly bind the given poses and obstacle-avoidance constraints. With multitask curriculum learning, SkyAgentX further improves timeline alignment and landmark consistency; its step count and turns more closely match the actual trajectory, terminal target identification is more accurate, and hallucinations are reduced. Nonetheless, its expression of geometric constraints such as altitude and speed remains insufficient, and some environmental elements are described redundantly.

\section{Discussion
}

\textbf{Scene Generalization Ability}. To investigate the generalization ability of various models across different embodied scenes, we assessed the performance of all models on four tasks corresponding to their respective scenes and compared the average BLEU scores across these tasks, as illustrated in Figure \ref{FIG:Analyze} (a). In the campus scene, the dense buildings and numerous obstacles, such as trees, generally lead to poor performance from all models. Conversely, in the residential area, which is smaller and contains fewer objects, all models demonstrate improved performance. Among the models, our proposed SkyAgent and SkyAgentX exhibit the best overall performance, significantly outperforming all other models in all four scenarios. In contrast, 3d-llm faces greater challenges due to the input being a three-dimensional scene rather than a modified image, resulting in subpar performance in each scenario.

\textbf{Task Generalization Ability}. To investigate the generalization ability of various models across different embodied tasks, we evaluate the performance of all models in four scenarios based on the tasks, comparing the average BLEU scores as depicted in Figure \ref{FIG:Analyze} (b). The results indicate that InstructBLIP and BLIP2 excel in Task $1$, which exclusively assesses the models’ captioning capabilities. In contrast, the Llava, MiniGPT, and MPLUG series models demonstrate superior performance in Task $4$, which necessitates the integration, comprehension, and response to information. Notably, our proposed SkyAgent and SkyAgentX models achieve state-of-the-art performance across all four tasks. They not only demonstrate exceptional captioning abilities in Task $1$ but also show superior capabilities in the more complex tasks (Tasks $2$, $3$, and $4$) that require reasoning, navigation, and planning, surpassing the other model series.

\textbf{The Impact of Scaling Law}. To examine the influence of model size on performance in embodied tasks, we selected three pairs of models with $7$ billion and $13$ billion parameters, comparing the average BLEU scores across the four scenarios within the four tasks, as illustrated in Figure \ref{FIG:Analyze} (c). The data reveal that while minor differences in performance arise due to varying model parameters, an increase in the number of parameters does not necessarily correlate with improved performance.

\textbf{Technology Ethics.} The potential risks of the autonomous drone technology involved in this study mainly include:

1. Privacy leakage risk: Visual sensors equipped on drones may inadvertently capture personal identification information (such as license plates or faces), requiring strict desensitization during the data collection phase. 

2. Airspace safety issues: If autonomous flight algorithms are exploited maliciously, they could lead to collision accidents. We have built no-fly zone settings into AeroSimulator (drones must not fly below 30m or above 50m).

\textbf{Scalability and System Requirements.} To assess the scalability of our system across diverse and complex 3D urban environments, we conduct experiments under different hardware configurations. The platform supports the loading of large-scale realistic 3D city scenes, with its upper bound determined primarily by available GPU memory and computational power. For instance, the Shenzhen scene can be smoothly rendered on a GPU with 6GB VRAM (e.g., NVIDIA GTX 1060), while other more complex scenes require higher-end hardware such as an NVIDIA RTX 3090 for stable operation. These results demonstrate the system’s flexibility and adaptability to varying performance constraints.

% 正文中是否要提到Openfly
\textbf{Novelty and Comparative Analysis.} This work’s primary novelty lies in establishing AeroVerse as the first comprehensive benchmark suite for aerospace embodied intelligence, addressing a critical gap in the literature. While powerful simulation tools like AirSim exist, they function as underlying engines rather than integrated benchmarks. The core research gap has been the absence of a full-stack platform that includes standardized task definitions, large-scale and domain-specific datasets, and unified evaluation protocols. AeroVerse is designed to fill this void. In comparison to recent related work such as OpenFly, AeroVerse offers several distinct contributions. Firstly, it is made public nearly six months prior, establishing its temporal precedence. Secondly, its scope is broader; whereas OpenFly concentrates on vision-language navigation, AeroVerse defines a five-dimensional task framework encompassing scene awareness, spatial reasoning, navigation, task planning, and motion decision. Thirdly, it introduces innovative data types, including the first large-scale real-world image-text pre-training dataset from a UAV perspective (AerialAgent-Ego15k), a virtual image-text-pose alignment dataset (CyberAgent-Ego500k), and five expert-labelled high-quality instruction datasets for downstream aerospace embodied tasks, which are crucial for pre-training and finetuning robust aerospace world models. Additionally, AeroVerse contains SkyAgentX, which pioneers as the first UAV-agent embodied large model,  unifying ``perception-reasoning-navigating-planning"  into an end-to-end framework by integrating aerospace embodied chain-of-thought and multitask curriculum learning. Finally, our benchmark provides key scientific insights by systematically evaluating current large 2D/3D visual-language models, revealing their common failure modes in complex aerospace embodied tasks and highlighting the need for specialized aerospace embodied world models.

\textbf{Bridging Simulation to the Real World.} A core design principle of AeroVerse is to facilitate the transfer of models from simulation to real-world applications (i.e., Sim-to-Real). The benchmark bridges this gap by integrating a large-scale real-world dataset (AerialAgent-Ego15k) with a high-fidelity simulated dataset (CyberAgent-Ego500k). This dual-pronged approach provides models with a training foundation that combines real-world perceptual capabilities with the diverse interactive learning opportunities available in simulation. To validate this approach, we have conducted preliminary real-world deployments. An aerospace embodied world model trained on the AeroVerse benchmark is successfully deployed on a physical UAV, achieving autonomous perception and navigation in real urban scenarios. This result provides initial evidence that models trained within our benchmark possess the potential to transfer effectively to real-world applications and can serve as a strong foundation for future development in areas such as urban inspection and logistics delivery.

\textbf{Limitations and Future Work.} While this study makes significant contributions, we acknowledge several limitations that pave the way for future research.

1. Regarding the generalization of 3D models, our findings indicate that existing models like 3D-LLM exhibit weak generalization in complex urban scenes, largely because they are primarily trained on indoor data and struggle with analyzing complex outdoor spatial relationships. Future work should focus on two key directions: 1) Domain Knowledge Injection, which involves encoding prior knowledge from urban planning (e.g., building height distributions, road hierarchies) as constraints to guide model learning, and 2) Enhanced Spatial Relationship Pretraining through dedicated pretraining phases to deepen the model's understanding of object relations in 3D space.

2. The diversity of simulation scenes is currently limited to four urban environments. Although representative, future iterations of AeroSimulator should incorporate more varied geographical settings, such as rural, mountainous, and coastal areas, to improve model robustness. We intend to integrate procedural generation techniques to dynamically create a wider range of scenes. We have already expanded our platform to include 20 high-fidelity 3D real-world urban scenes, significantly increasing the benchmark's richness and the potential for model generalization.

3. The scale of real-world data, while pioneering, can be expanded. Although AerialAgent-Ego15k is the first large-scale dataset of its kind, we plan to continuously augment it with more data covering diverse weather conditions, lighting scenarios, and dynamic events to further close the Sim-to-Real gap.

\section{Conclusion}
This paper addresses the research gap in aerospace embodied world models by developing AeroVerse, a comprehensive benchmark suite designed to empower UAV intelligent agents with end-to-end autonomous capabilities. We establish the AeroSimulator platform with four realistic urban scenarios, introduced two pioneering pre-training datasets (the real-world AerialAgent-Ego15k and the virtual CyberAgent-Ego500k), clearly delineated five standardized downstream tasks for the first time, and constructed corresponding instruction fine-tuning datasets. Our extensive experiments systematically reveal, for the first time, the widespread limitations of current mainstream 2D and 3D visual-language models in aerospace embodied tasks, highlighting critical areas for improvement such as spatial reasoning and long-term planning. Furthermore, we propose SkyAgentX, the first UAV-agent embodied large model integrating ``perception-reasoning-navigating-planning", which demonstrates superior performance, thus validating the effectiveness of our benchmark and providing a strong baseline for future models.

The significance of this study lies in providing the research community with a standardized platform for fair comparison and rapid iteration. The release of AeroVerse, along with our findings, will direct stakeholders and researchers by offering valuable references and a clear pathway for developing more robust, capable, and safer aerospace intelligent agents. In the future, we plan to expand the simulation environments and continue to refine the training datasets and downstream tasks, promoting applications in areas such as river garbage detection, power inspection, and forest fire rescue, thereby unlocking the full application value of aerospace embodied intelligence.

\bibliography{ref}
\bibliographystyle{IEEEtran}

\begin{IEEEbiography}[{\includegraphics[width=1in,height=1.25in,clip,keepaspectratio]{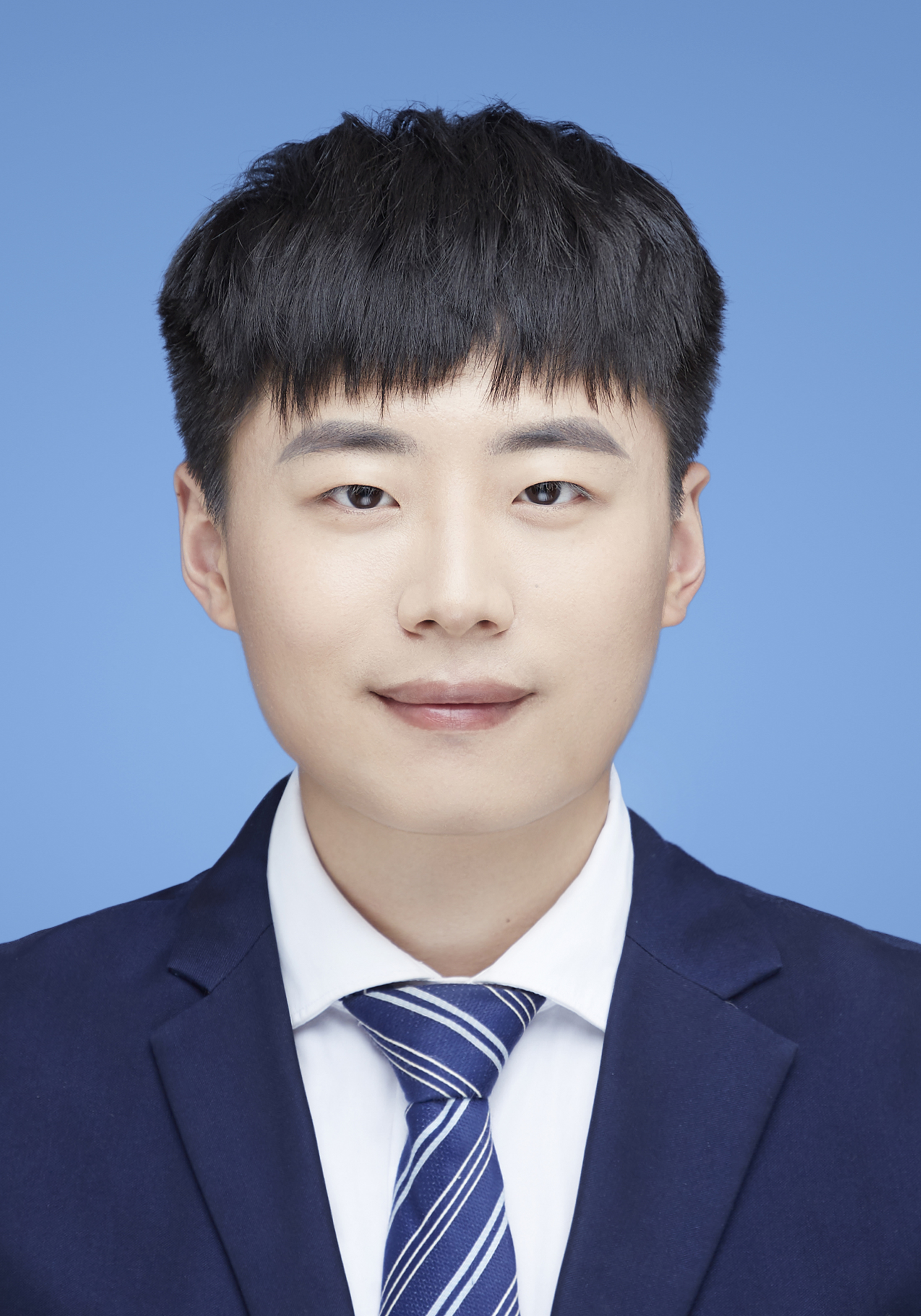}}]{Fanglong Yao}
        received the B.Sc. degree from Inner Mongolia University, Hohhot, China, in 2017, and the Ph.D. degree from the Aerospace Information Research Institute, Chinese Academy of Sciences, Beijing, China, in 2022. He is currently an associate professor with the Aerospace Information Research Institute, Chinese Academy of Sciences. 

        His research interests include spatial intelligence, embodied intelligence, and swarm intelligence, concentrating on multi-agent learning, multimodal fusion, 3D scene understanding and spatiotemporal data analysis. 
	\end{IEEEbiography}
 \vspace{-2cm}
 \begin{IEEEbiography}[{\includegraphics[width=1in,height=1.25in,clip,keepaspectratio]{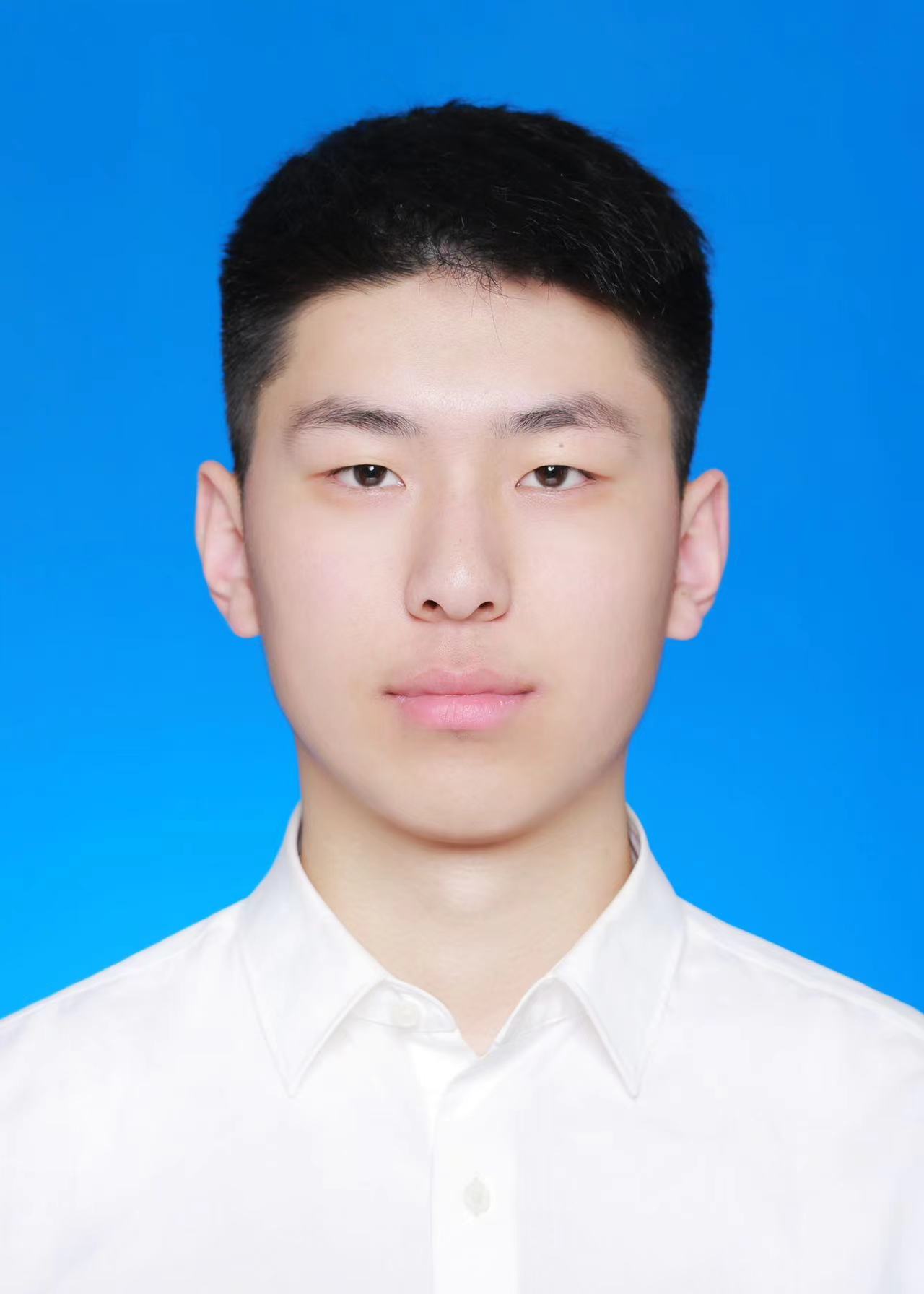}}]{Yuanchang Yue}
        received the B.Sc. degree from jiangnan University, wuxi, China, in 2022. He is currently a master's student with the Aerospace Information Research Institute, Chinese Academy of Sciences. 

        His research interests include embodied intelligence, and task planning. 
	\end{IEEEbiography}
\vspace{-2cm}
  \begin{IEEEbiography}[{\includegraphics[width=1in,height=1.25in,clip,keepaspectratio]{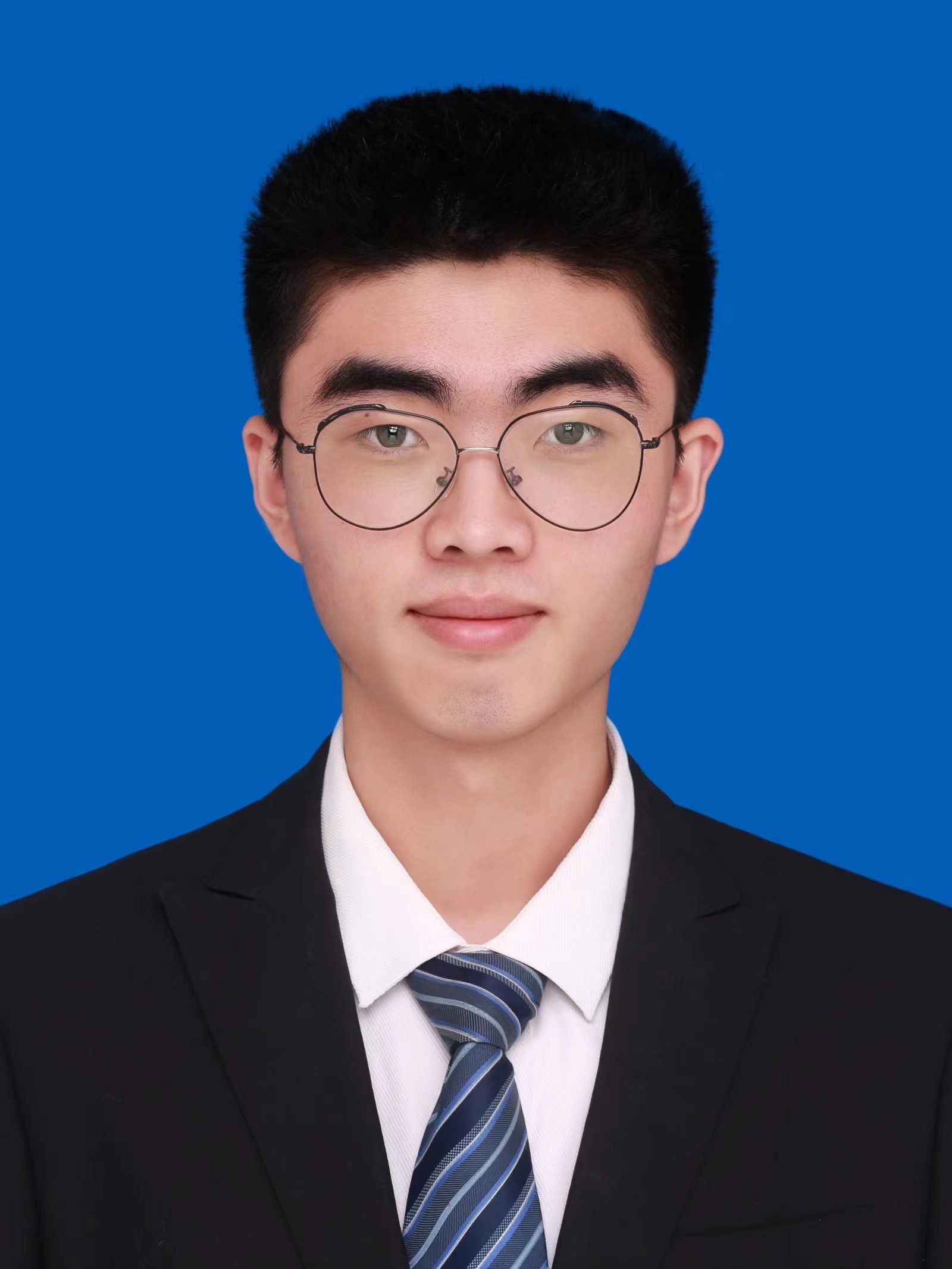}}]{Youzhi Liu}
        received the B.Sc. degree from Hunan University, changsha, China, in 2022. He is currently a Ph.D student with the Aerospace Information Research Institute, Chinese Academy of Sciences. 

        His research interests include embodied intelligence, and visual language navigation. 
	\end{IEEEbiography}
 \vspace{-1.5cm}

 	\begin{IEEEbiography}[{\includegraphics[width=1in,height=1.25in,clip,keepaspectratio]{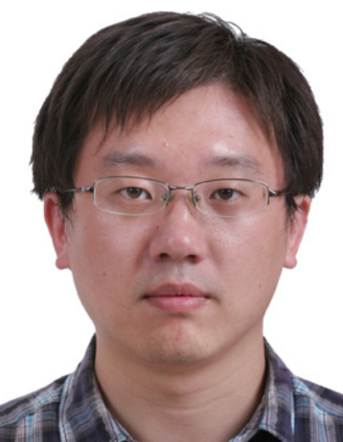}}]{Xian Sun}
		received the B.Sc. degree from the Beijing University of Aeronautics and Astronautics, Beijing, China, in 2004, and the M.Sc. and Ph.D. degrees from the Institute of Electronics, Chinese Academy of Sciences, Beijing, in 2009. 
		
		He is a Professor with the Aerospace Information Research Institute, Chinese Academy of Sciences, Beijing, China. His research interests include computer vision, geospatial data mining, and remote sensing image understanding.
	\end{IEEEbiography}
	\begin{IEEEbiography}[{\includegraphics[width=1in,height=1.25in,clip,keepaspectratio]{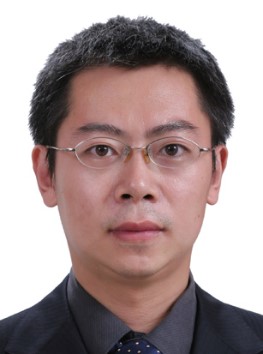}}]{Kun Fu}
		 received the B.Sc., M.Sc., and Ph.D. degrees from the National University of Defense Technology, Changsha, China, in 1995, 1999, and 2002, respectively. 
		 
		 He is a Professor with the Aerospace Information Research Institute, Chinese Academy of Sciences, Beijing, China. His research interests include computer vision, remote sensing image understanding, geospatial data mining, and visualization.
	\end{IEEEbiography}

\newpage

% \end{IEEEbiography}

\appendix

%\vfill

\end{document}